\documentclass{article}

\PassOptionsToPackage{numbers, compress}{natbib}
\usepackage[main, preprint]{neurips_2026}

\usepackage[utf8]{inputenc} % allow utf-8 input
\usepackage[T1]{fontenc}    % use 8-bit T1 fonts
\usepackage[colorlinks=true]{hyperref}       % hyperlinks
\usepackage{url}            % simple URL typesetting
\usepackage{booktabs}       % professional-quality tables
\usepackage{amsfonts}       % blackboard math symbols
\usepackage{nicefrac}       % compact symbols for 1/2, etc.
\usepackage{microtype}      % microtypography
\usepackage{xcolor}         % colors

\usepackage{amsmath}
\usepackage
{
    amsmath,
    amssymb,
    amsthm,
    dsfont, % for 1 vector or indicator function
    enumitem,
    graphicx,
    mathtools,
    nicefrac,
    tabularx,
    tcolorbox,
    tikz,
    xcolor,
}

\newcommand{\R}{\mathbb{R}}                                      % real numbers
\newcommand{\ts}{\hspace*{0.1em}}                                % thin space
\newcolumntype{C}[1]{>{\centering\let\newline\\\arraybackslash\hspace{0pt}}m{#1}}

\DeclareMathOperator{\tr}{tr}
\DeclareMathOperator{\mspan}{span}

\newtheorem{theorem}{Theorem}[section]

\newtheorem{definition}[theorem]{Definition}

\newtheorem{remark}[theorem]{Remark}

\usepackage{graphicx}
\usepackage[scr=boondoxupr]{mathalfa}
\usepackage{algpseudocode}
\usepackage{algorithm}
\usepackage{subcaption}

\graphicspath{{pics/}}

\usepackage[justification=justified,format=plain]{caption}

\usepackage[labelfont=bf, position=top]{caption}

\usetikzlibrary{arrows, arrows.meta, shapes, fit, automata, positioning}

\title{How deep is your network? \\ Deep vs.\ shallow learning of transfer operators}

\author{%
  Mohammad Tabish \\
  Maxwell Institute for Mathematical Sciences, \\
  University of Edinburgh and Heriot--Watt University, \\
  Edinburgh, UK \\
  \texttt{M.Tabish-1@sms.ed.ac.uk}
  \And
  Benedict Leimkuhler \\
  School of Mathematics, \\ University of Edinburgh, \\
  Edinburgh, UK \\
  \texttt{B.Leimkuhler@ed.ac.uk}
  \And
  Stefan Klus \\
  School of Mathematical and Computer Sciences,\\ Heriot--Watt University, \\
  Edinburgh, UK \\
  \texttt{S.Klus@hw.ac.uk}
}

\date{}

\begin{document}

\maketitle

\begin{abstract}

We introduce RaNNDy, a randomized neural network framework for learning transfer operators and their spectral decompositions. By fixing the weights of the hidden layers and training only the output layer, RaNNDy achieves significant reductions in computational costs---often by two orders of magnitude---with similar accuracy compared to state-of-the-art deep learning methods. The method is less sensitive to hyperparameter tuning and provides a closed-form solution for the output layer weights that directly represent eigenfunctions of the learned operator. Moreover, it is possible to estimate uncertainties associated with the computed spectral properties via ensemble learning. We demonstrate the efficacy of the proposed approach using diverse benchmark problems, including stochastic dynamical systems and protein folding processes, highlighting the strengths but also weaknesses of RaNNDy.

\end{abstract}

\section{Introduction}
\label{sec:introduction}

Data-driven approximations of operators describing, for example, the evolution of probability densities, observables, or wave functions are of key importance in applied mathematics and machine learning. Although such operators are typically infinite-dimensional, their finite-dimensional approximations can help us analyze systems that might be challenging to model or understand. Transfer operators such as the Koopman operator and the Perron--Frobenius operator, for instance, can be used to gain crucial insights into the evolution of complex dynamical systems, e.g., protein folding processes or fluid flows. Spectral decompositions of these operators allow us to analyze the global behavior of complex systems, e.g., metastable states in molecular dynamics or coherent sets in fluid dynamics, without requiring detailed mathematical models \cite{klus2024dynamical}. For an introduction to transfer operator theory, we refer to \cite{dellnitz1999approximation, mezic2005spectral, lasota2013chaos, KKS16}. Data-driven algorithms for approximating transfer operators serve as a bridge between theoretical models and practical computations.

\paragraph{Related works:} The most popular methods for approximating transfer operators are \emph{Ulam's method}~\cite{Ulam60}, \emph{extended dynamic mode decomposition} (EDMD) \cite{williams2015data, KKS16, TABISH2025134822}, kernel-based variants such as \emph{kernel EDMD} (kEDMD) \cite{WRK15, KSM20}, as well as \emph{generator EDMD} (gEDMD) \cite{KNPNCS20}. Most of these techniques require a function space spanned by a set of fixed basis functions or a reproducing kernel Hilbert space spanned by a given kernel to approximate the operator. The accuracy of the approximation depends strongly on the choice of this function space. Neural network-based dictionary learning methods have been successfully used for approximating transfer operators in combination with EDMD see, e.g., \cite{li2017extended, enoch2019, gulina2021two, xiao2022deep}, and have been applied to various domains such as modeling and control of autonomous vehicles \cite{xiao2022deep} and protein folding \cite{mardt2018vampnets}. VAMPnets \cite{mardt2018vampnets} use the output layer of the network as a set of basis functions. The network is trained by optimizing the VAMP-2 score \cite{wu2020variational}, a loss function based on the variational principle for transfer operators~\cite{noe2013variational}. Typically, training these neural networks requires backpropagation that iteratively updates the parameters (weights and biases) of the network. As the number of neurons and hidden layers increases, the computational complexity of the training algorithm grows significantly, leading to problems such as slow convergence, the existence of numerous local minima, exploding/vanishing gradients, and sensitivity to hyperparameters such as the learning rate~\cite{gori1992problem, TEBRAAKE199571}. To address these challenges, more sophisticated approaches have been proposed, e.g., adaptive methods like Adam \cite{staib2019escaping, kingma2014adam} for faster training and escaping saddle points, and gradient clipping for exploding/vanishing gradients \cite{pascanu2013difficulty}.

Neural networks with random weights \cite{zhang2016survey, suganthan2021origins}, on the other hand, aim to avoid or at least mitigate some of the aforementioned problems. In such networks, the weights of the hidden layers are randomly selected and kept fixed throughout the training and only the output layer of the network is trained with either a closed-form solution or iteratively, see \cite{malik2023random, CAO2018278, ZHANG2016146} for an overview. This makes the architecture simple and easy to train. Single-layer feedforward neural network architectures with random weights, such as random vector functional link (RVFL) networks \cite{pao1992functional}, extreme learning machines (ELM) \cite{HUANG2006489}, and broad learning systems (BLS) \cite{chen2017broad}, have been studied and utilized for various classification and regression tasks and have also been shown to have the universal approximation property \cite{park1991universal, scarselli1998universal, chen2018universal}. The above characteristics make them a powerful alternative for approximating operators. Random feature models have been used for partial differential equations (PDEs) \cite{nelsen_2021} and developed further by designing function-valued random features. For further details on neural operator learning for PDEs, their connections to kernel methods, and computational and theoretical advantages of random feature models, see \cite{nelsen_2024} and references therein. Moreover, \cite{bolager2024koopman} introduces a Koopman-informed method to construct the optimal hidden layer weights for recurrent neural networks without using gradient descent and presents applications to forecasting chaotic dynamical systems with comparable accuracy and reduced training time.

\paragraph{Contributions:} We propose a novel data-driven framework, called RaNNDy, based on randomized neural networks and variational principles, to approximate different operators such as the Koopman operator and the Perron--Frobenius operator. We show how the randomization allows us to compute a closed-form solution for the output layer, representing the eigenfunctions of the operator using more interpretable basis functions. We demonstrate that one of the main advantages is that we can approximate the eigenfunctions of different operators more efficiently, while at the same time avoiding common problems associated with deep learning. Furthermore, we show that RaNNDy enables us to quantify the uncertainties in the data-driven approximation of transfer operators via ensemble learning. We illustrate the advantages of the framework with the aid of numerical examples ranging from the overdamped Langevin equation to high-dimensional protein folding processes.

\paragraph{Structure of the paper: }We first discuss the data-driven approximation of self-adjoint and non-self-adjoint operators using variational principles as well as transfer operators for reversible and non-reversible systems in Section~\ref{sec:operator_approximation}. In Section~\ref{sec:ranndy_operator_approximation}, we describe the proposed data-driven framework for the approximation of linear operators using randomized neural networks. We present numerical results for different operators and systems in Section~\ref{sec:numerical_experiments} and highlight open problems and future research directions in Section~\ref{sec:conclusion}.

\section{Data-driven approximation of operators}
\label{sec:operator_approximation}

% We will start by briefly discussing data-driven approximations of spectral properties of operators using variational principles as well as transfer operators for reversible and non-reversible systems. %The notation used throughout the manuscript is summarized as follows: $X_t$ denotes a stochastic process, $p_\tau$ denotes the transition density functions for a fixed lag time $\tau$, the Koopman, Perron--Frobenius, and forward-backward operators are denoted by $\mathcal{K}^\tau$ $\mathcal{P}^\tau$, and $\mathcal{F}^\tau$, respectively, $\pi$ is the stationary distribution for the process. Furthermore, $\mathcal{A}$ denotes a general operator and $\mathcal{A}_{\psi}$ its finite-dimensional approximation. $A$ is the matrix representation of $\mathcal{A}_{\psi}$.

\subsection{Self-adjoint operators}

In what follows, let $H$ be a Hilbert space and $\mathcal{A}\colon H \to H$ a compact self-adjoint linear operator, i.e., $\langle\mathcal{A} f, g \rangle = \langle f, \mathcal{A} g \rangle$ for all $f, g \in \mathcal{D}(\mathcal{A})$, where $\mathcal{D}(\mathcal{A})$ is the domain of $\mathcal{A}$ and $\langle \cdot, \cdot \rangle$ the inner product associated with $H$.
\begin{definition}[Rank-one operator]
For $r, s \in H$, the \emph{rank-one operator} $s \otimes r\colon H \rightarrow H$ is defined by $(s \otimes r)f = \langle r, f\rangle s$.
\end{definition}
It is well-known that there exists a spectral decomposition of $\mathcal{A}$ given by $ \mathcal{A} = \sum_i \lambda_i (\varphi_i \otimes \varphi_i)$
where $\{\varphi_i\}_i$ are orthonormal eigenfunctions corresponding to the eigenvalues $\{\lambda_i\}_i \subset \R $. Assuming that the eigenvalues $\lambda_i$ are sorted in non-increasing order, i.e., $\lambda_1 \geq \lambda_2 \geq \lambda_3 \geq \dots$, we can compute eigenfunctions using variational formulations \cite{eschwe2004variational, nuske2014variational}.

\begin{theorem}\label{thoerem:variational_principle_eigv}
Assume that $\hat{\varphi}_i$ is orthogonal to the previous $i - 1$ eigenfunctions, then it holds that
\begin{align*}
    \max_{\hat{\varphi}_i}
    \frac{\langle \hat{\varphi}_i, \mathcal{A} \hat{\varphi}_i \rangle}{\langle \hat{\varphi}_i, \hat{\varphi}_i \rangle} = \lambda_i
\end{align*}
and the Rayleigh quotient is maximized by the $ i $th eigenfunction.
\end{theorem}
In Appendix \ref{appendix:approx_self_adjoint}, we show how a matrix representation of the operator $ \mathcal{A} $, restricted to finite-dimensional subspace, can be computed from data using the above variational principle.

\paragraph{Transfer operators for reversible systems: } Transfer operators associated with reversible stochastic processes, which play an important role in molecular dynamics, are self-adjoint. Consider a stochastic process $\{\mathbf{X}_t\}$ defined on a state space $\mathbb{X} \subset \mathbb{R}^d$.
%and suppose that it is governed by a \emph{stochastic differential equation} (SDE) of the form
% \begin{equation}\label{eq:sde}
%     \mathrm{d}X_t = b(X_t) \ts \mathrm{d}t + \sigma(X_t) \ts \mathrm{d}W_t,
% \end{equation}
% where $ b \colon \mathbb{R}^d \rightarrow \mathbb{R}^d $ is the drift term, $\sigma \colon \mathbb{R}^d \rightarrow \mathbb{R}^{d \times d}$ is the diffusion term, and $ W_t $ is a $ d $-dimensional Wiener process.
Let $p_{\tau}\colon \mathbb{X} \times \mathbb{X} \rightarrow \mathbb{R}$ be the transition density function, i.e., $p_{\tau}(x, y)$ is the probability of $\mathbf{X}_{t+\tau} = y$ given $\mathbf{X}_t = x$, where $\tau > 0 $ is a fixed lag time. Stochastic processes can be analyzed using transfer operators such as the Perron--Frobenius operator and the Koopman operator, which describe the evolution of probability densities and observables, respectively. In what follows, let $L^r(\mathbb{X})$ denote the space of $r$-Lebesgue integrable functions for $1 \leq r \leq \infty$. The \emph{Perron--Frobenius operator} $\mathcal{P}^\tau$ is defined by
\begin{align*}
    \mathcal{P}^\tau p(x) = \int_{\mathbb{X}} p_{\tau}(y, x) \ts p(y) \ts \mathrm{d}y.
\end{align*}
The \emph{Koopman operator} $\mathcal{K}^\tau$ is defined by
\begin{align*}
    \mathcal{K}^\tau f(x) = \int_{\mathbb{X}} p_{\tau}(x, y)f(y) \ts \mathrm{d}y = \mathbb{E}[f(\mathbf{X}_{t+\tau}) \mid \mathbf{X}_t=x].
\end{align*}
The Perron--Frobenius operator determines the probability density, and its adjoint, the Koopman operator, the expected value of the observable after time $\tau$. We are interested in eigenvalues $ \lambda_i $ and the corresponding eigenfunctions $ \varphi_i $ of these transfer operators since they help us identify, for example, metastable states in molecular dynamics and coherent sets in fluid flows \cite{mezic2005spectral, schutte2013metastability, KKS16, banisch2017understanding}. We focus in particular on the Koopman operator, i.e., $\mathcal{K}^\tau \varphi_i = \lambda_i(\tau) \ts \varphi_i$.

\begin{definition}[Invariant density]
The density $\pi(x)$ is called stationary (or invariant) if it satisfies $\mathcal{P}\pi = \pi$, i.e., $\pi$ is an eigenfunction of $\mathcal{P}$ corresponding to the eigenvalue $\lambda = 1$.
\end{definition}

If the invariant distribution $\pi$ exists, the above operators are well-defined as $\mathcal{P}^\tau\colon L^2_{1/\pi}(\mathbb{X}) \rightarrow L^2_{1/\pi}(\mathbb{X})$ and $\mathcal{K}^\tau\colon L^2_{\pi}(\mathbb{X}) \rightarrow L^2_{\pi}(\mathbb{X})$. For details on the domains of the above operators, see \cite{koltai2018optimal}.

\begin{definition}[Detailed balance]
The process $\mathbf{X}_t$ is called \emph{reversible} if the detailed balance condition is satisfied, i.e., $\forall x, y \in \mathbb{X}$, we have $ \pi(x) \ts p_\tau (x,y) = \pi(y) \ts p_\tau(y, x)$.
\end{definition}

It is important to note that detailed balance implies that the Koopman operator $\mathcal{K}^\tau$ and the Perron--Frobenius operator $\mathcal{P}^\tau$ are self-adjoint (with respect to suitably reweighted inner products). Hence, the eigenvalues $\lambda_i$ of the Koopman operator are real-valued, and the eigenfunctions $\varphi_i$ form an orthogonal basis with respect to the stationary density $\langle \cdot, \cdot \rangle_\pi$ \cite{klus2018data}. Thus, we can expand a function $f \in L^2_\pi(\mathbb{X})$ in terms of eigenfunctions, i.e., $f = \sum_{i=1}^\infty \langle f, \varphi_i \rangle_\pi \ts \varphi_i$ such that
$\mathcal{K}^\tau f = \sum_{i=1}^\infty \lambda_i(\tau) \langle f, \varphi_i \rangle_\pi \ts \varphi_i$.
Furthermore, the eigenvalues decay exponentially as $\lambda_i(\tau) = e^{-\kappa_i \tau}$, where $\kappa_i$ is the relaxation rate~\cite{noe2013variational}. In particular, we have
$1 = \lambda_1 > \lambda_2 \geq \lambda_3 \geq \dots$, indicating that, for a sufficiently large $\tau$, the first few dominant eigenvalues explain the dynamics well.

\subsection{Non-self-adjoint operators}

For a non-self-adjoint compact linear operator $\mathcal{A}\colon H_1 \rightarrow H_2$, where $H_1$ and $H_2$ are Hilbert spaces with associated inner products $\langle \cdot, \cdot\rangle_{H_1}$ and $\langle \cdot, \cdot\rangle_{H_2}$, there exists a singular value decomposition (SVD) of the form $\mathcal{A} = \sum_i \sigma_i (\chi_i \otimes \eta_i)$
Here, $\{\sigma_i\}_i \subset \mathbb{R}_{>0}$ are the nonzero singular values, sorted in non-increasing order so that $\sigma_1 \ge \sigma_2 \ge \sigma_3 \ge \dots$, and the corresponding left singular functions $\{\chi_i\}_i \subset H_2$ and right singular functions $\{\eta_i\}_i \subset H_1$ are orthonormal \cite{MSKS20}. If there are infinitely many nonzero singular values, they form a null sequence. Assuming that $ f $ is orthogonal to the first $i-1$ right singular functions $ \eta_i $ and $ g $ is orthogonal to the first $i-1$ left singular functions $\chi_i$, it holds that
\begin{align*}
    \max_{f, g} \frac{\langle g, \mathcal{A} f \rangle_{H_2}}{\sqrt{\langle f, f \rangle_{H_1} \langle g, g \rangle_{H_2}}} = \sigma_i.
\end{align*}
This is consistent with the variational principle for the eigenvalues and eigenfunctions and provides a way to formulate an optimization problem to approximate the singular values and singular functions of non-self-adjoint operators. In Appendix \ref{appendix:approx_non_self_adjoint}, we derive matrix representations of finite-dimensional approximations of non-self-adjoint operators $\mathcal{A}$.

\paragraph{Transfer operators for non-reversible systems: } For non-reversible dynamical systems, associated transfer operators are in general not self-adjoint. That is, their eigenvalues and eigenfunctions may be complex-valued. In this case, we approximate the singular values and singular functions of these operators to analyze the system.
% \begin{align*}
%     \mathcal{T}^\tau u(x) = \frac{1}{\nu(x)} \int_{\mathbb{X}} p_{\tau}(y, x) u(y) \mu(y) \ts \mathrm{d}y,
% \end{align*}
%     where $\nu = \mathcal{P}\mu$. %Furthermore, let $\mathcal{K}^\tau \colon L^2_\nu \rightarrow L^2_\mu$ be the adjoint of $\mathcal{T}^\tau$, given by
    % \begin{align*}
    %     \mathcal{K}^\tau f(x) = \int_{\mathbb{X}} p_{\tau}(x, y) \ts f(y) \ts \mathrm{d}y.
    % \end{align*}
% The dominant singular functions of these operators help us analyze non-reversible systems.
In fluid dynamics, these singular functions have been used to identify coherent sets~\cite{banisch2017understanding}. For more details, we refer to \cite{koltai2018optimal, klus2019kernel, wu2020variational}. The Perron--Frobenius operator reweighted with respect to a density $\mu$, denoted by $\mathcal{T}^\tau \colon L^2_\mu \rightarrow L^2_\nu$, is defined as $\mathcal{T}^{\ts\tau} \ts u(x) = \frac{1}{\nu(x)} \int_{\mathbb{X}} p_{\tau}(y, x) u(y) \mu(y) \ts \mathrm{d}y$, where $\nu = \mathcal{P}^\tau\mu$. It propagates densities w.r.t.\ the reference density $\mu$ and its adjoint the Koopman operator $\mathcal{K}^\tau \colon L^2_\nu \rightarrow L^2_\mu$ propagates the observables, where $\nu = \mathcal{P}\mu$.

\section{Operator approximation using randomized neural networks}
\label{sec:ranndy_operator_approximation}

In randomized neural networks, the weights of the hidden layers are randomly selected and kept fixed throughout training. Only the output layer of the network is trained, either by computing a closed-form solution or using iterative methods, see \cite{malik2023random, CAO2018278, ZHANG2016146} for an overview. This avoids backpropagation and makes the architecture simple and easy to train. The input data are transformed into a random feature space with hidden layers acting as a \emph{random feature map} (RFM). The output layer then represents the output of the network.
% In order to classify data into $c$ classes, let $X = [x_1, x_2, \dots, x_m] \in \mathbb{R}^{d \times m}$ be the training data matrix containing $m$ data points of dimension $d$ and $Y = [y_1, y_2, \dots, y_m] \in \mathbb{R}^{c \times m}$ the matrix containing the corresponding labels in the one-hot encoding format. Then
The model is a function $\textbf{f}: \mathbb{R}^d \rightarrow \mathbb{R}^n$, defined by
\begin{align*}
    \textbf{f}(x) = W_o \cdot \sigma(W x + b),
\end{align*}
where $\sigma$ is the activation function, $W \in \mathbb{R}^{N \times d}$ are the weights of the hidden layer, $b \in \mathbb{R}^N$ is the bias term, and $W_o \in \mathbb{R}^{n \times N}$ is the matrix containing the output weights. The RFM is $R(x) = \sigma(Wx + b)$.

\subsection{Learning linear operators}
We will now introduce the proposed framework for approximating a compact linear operator $\mathcal{A}$ and its spectral properties using randomized neural networks. Of particular interest is the approximation of the eigenvalues $ \lambda_i $ and the associated eigenfunctions $ \varphi_i $ satisfying $ \mathcal{A} \varphi_i = \lambda_i \ts \varphi_i $ and the approximation of singular values $ \sigma_i $ and associated left and right singular functions $ \chi_i $ and $ \eta_i $ such that $\mathcal{A} \eta_i = \sigma_i \ts \chi_i $ and $ \mathcal{A}^* \chi_i = \sigma_i \ts \eta_i$. Considering the strength of neural networks to approximate nonlinear functions from data, we design a randomized neural network called RaNNDy with two variants. In the first variant, the output layer represents the optimal basis functions. The network generates a high-dimensional random feature space using the random feature map (RFM) of the hidden layers. The data is transformed through the RFM and passed to the output layer that represents the basis functions. The training of the parameters in the output layer is done iteratively via loss functions formulated using the variational principle discussed in the previous section. In the second variant, taking advantage of randomization in the network, we formulate an optimization problem to directly approximate the eigenfunctions represented by the output layer of the network. In this case, the RFM acts as a randomized basis, see the figure in Appendix \ref{appendix:ranndy_variants}.

%shows the two variants of the network. The first one is the optimal basis approximator, which is similar to VAMPnets as it outputs the basis functions for the operator approximation and is trained iteratively. The second one is the eigenfunction approximator, in which the output layer represents the eigenfunctions of the operator, computed directly from an eigenvalue problem, i.e., we have a closed-form solution for the output layer.

\subsection{Data and iterative training of the output layer}

To approximate the operator $\mathcal{A}$, we need $m$ training data points $\big( \psi(x_i), \mathcal{A} \psi(x_i) \big)_{i=1}^m$. For the Koopman operator, let $x_i, y_i \in \mathbb{X}$ denote the data points from a dynamical system with $x_i = \mathbf{X}_t$ and $y_i = \mathbf{X}_{t+\tau}$. Data can be obtained by simulating the dynamical system or by repeatedly measuring its state to obtain the matrices $ X = [x_1, x_2, \dots, x_m] \quad \text{and} \quad Y = [y_1, y_2, \dots, y_m] \in \mathbb{R}^{d \times m}$ containing $m$ data points (or snapshots) of a $d$-dimensional system, where $Y$ contains the states of the system after lag time $\tau$. Hence, our training data is given by $ \big( \psi(x_i), \mathcal{K}^\tau\psi(x_i) \big)_{i=1}^m = \big( \psi(x_i), \psi(y_i) \big)_{i=1}^m$.

\paragraph{Basis functions via RaNNDy: } For the sake of simplicity, we derive the framework for networks with a single hidden layer. However, multiple hidden layers will follow similarly. The $n$ neurons in the output layer represent the $n$ basis functions $\psi_i \in \mathcal{D}(\mathcal{A})$, with $i = 1, 2, \dots, n$, to approximate the operator $\mathcal{A}$. For the $m$ data points contained in $X \in \mathbb{R}^{d \times m}$, the randomized features are represented by $ R(X) \in \mathbb{R}^{N \times m}$. The output layer of the network will then be the map $\psi \colon \R^N \rightarrow \R^n$ given by $\psi(x, W_o) = \sigma(W_o R(x))$. Hence, for fixed network weights, after passing the data matrix $X$ through the network, we get the transformed data matrix $\Psi_0 = \psi(X, W_o) \in \mathbb{R}^{n \times m}$. In order to train the output layer weights $W_o$ of the network, we use the variational principle discussed above. For convenience, let us denote $\psi(\ts \cdot \ts, W_o)$ by $\psi$. Consider the initial output layer weights, $W_o \in \mathbb{R}^{n \times N}$, which define the basis $\psi = \sigma(W_o R(\ts \cdot \ts))$. Then, using the variational principle in Appendix~\ref{appendix:approx_self_adjoint} for a function $f = \sum_{i=1}^n w_i \psi_i$, we obtain the Rayleigh quotient
\begin{align*}
    \max_f \frac{\langle f, \mathcal{A} f \rangle}{\langle f, f \rangle} = \underset{\substack{w}}{\max} \frac{w^\top C_{01}(W_o) w}{w^\top C_{00}(W_o) w}.
\end{align*}
Since the basis is represented by the output layer, the matrices $C_{01}(W_o) \in \mathbb{R}^{n \times n}$ and $C_{00}(W_o) \in \mathbb{R}^{n \times n}$ now depend on $ W_o $. The above quotient leads to the eigenvalue problem $A(W_o) \ts w = \lambda \ts w$, where for a given output layer $W_o$, the matrix $A(W_o) = C_{00}^{-1}(W_o) \ts C_{01}(W_o)$ represents the projected operator $\mathcal{A}_{\psi}$. The solutions of this eigenvalue problem are orthonormal vectors with respect to the inner product weighted by the matrix $C_{00}(W_o)$ \cite{nuske2014variational}. The eigenvectors $w^{(i)}$ of the matrix $A(W_o)$ can be used to construct the eigenfunctions of the projected operator $\mathcal{A}_{\psi}$ as shown in Appendix \ref{appendix:approx_self_adjoint}. Furthermore, using the variational principle, the sum of the eigenvalues, i.e., $\sum_{i=1}^n \lambda_i = \max_{\psi_1, \dots, \psi_n} \sum_{i=1}^n \langle \psi_i, \mathcal{A} \psi_i \rangle, $ s.t.\ $ \langle \psi_i, \psi_j \rangle = \delta_{ij}$,
can be optimized to get the best approximation of the eigenfunctions from data. Hence, the sum of eigenvalues of the estimated matrix $A(W_o)$ can be maximized (or minimized, depending on the operator) using, for example, gradient descent at each training epoch, which then gives the updated basis functions with the new weights, $W_o$, of the output layer. The data-driven approximations of $C_{00}(W_o)$ and $C_{01}(W_o)$ are
\begin{align*}
    \widehat{C}_{00}(W_o) &= \frac{1}{m}\sum_{i=1}^m \psi(x_i, W_o) \ts \psi(x_i, W_o)^\top = \frac{1}{m}\Psi_0(W_o) \ts \Psi_0(W_o)^\top, \\
    \widehat{C}_{01}(W_o) &= \frac{1}{m}\sum_{i=1}^m \psi(x_i, W_o)(\mathcal{A} \ts  \psi (x_i, W_o))^\top = \frac{1}{m}\Psi_0(W_o) \ts \Psi_1(W_o)^\top,
\end{align*}
where $\Psi_0(W_o) = \psi(X, W_o) \in \R^{n \times m}$ and $\Psi_1(W_o) = [(\mathcal{A}\psi)(x_1), (\mathcal{A}\psi)(x_2), \dots , (\mathcal{A}\psi)(x_m)] \in \R^{n \times m}$. With $\widehat{A}(W_o) = \widehat{C}_{00}(W_o)^{+} \widehat{C}_{01}(W_o)$, we thus get the loss function
\begin{align*}
    % \max_{W_o \in \mathbb{R}^{n \times N}} \tr\left(\widehat{C}_{00}(W_o)^{+} \widehat{C}_{01}(W_o)\right)
    % \iff
    \max_{W_o \in \mathbb{R}^{n \times N}} \tr\left(\widehat{A}(W_o)\right),
\end{align*}
which can be used to iteratively train the network, see also \cite{nuske2014variational, mardt2018vampnets}.

\subsection{Can we directly approximate eigenfunctions?}

The randomization in our network, i.e., defining the randomly sampled weights to be fixed, can lead to a straightforward optimization problem that allows us to compute a closed-form solution for the output layer. We leverage this fact to directly approximate the eigenfunctions of a self-adjoint operator $\mathcal{A}$, which will be represented by the output layer. In this case, the random features in the hidden layers will act as a randomized basis. Hence, we redefine the basis $\psi\colon \mathbb{R}^d \rightarrow \mathbb{R}^N$ by $ \psi(x) = R(x) = \sigma(Wx+b)$. That is, we now obtain matrices $C_{00}, C_{01} \in \R^{N \times N}$.
%In order to approximate the dominant eigenfunction of the self-adjoint operator $\mathcal{A}$ using the randomized basis, we can maximize
% \begin{align*}
% v = \underset{\substack{W_o \in \R^N}}{\operatorname{arg\,max}} \quad \frac{W_o^\top C_{01} W_o}
%      {W_o^\top C_{00} W_o},
% \end{align*}
% where $W_o \in \R^{N \times 1}$ is the vector of output layer weights for a single neuron output and $v$ is the dominant eigenvector of the matrix $A$ representing the projected operator $\mathcal{A}_{\psi}$. As shown in the previous section,
The eigenvalues of $A$ correspond to the different orthonormal eigenvectors with respect to the inner product weighted by $C_{00}$ and those eigenvectors lead to different orthonormal eigenfunctions. Let $W_o \in \mathbb{R}^{N \times n}$ be the matrix of the output layer weights for $n$ outputs. In order to obtain the top $n$ eigenvectors of the matrix $A$, we can optimize the trace of $ W_o^\top C_{01} W_o $ with vectors in $W_o$ being orthonormal, i.e., $W_o^\top C_{00} W_o = I$, where $I$ is the identity matrix of size $n$. This results in the optimization problem
\begin{align*}
    \max_{W_o \in \mathbb{R}^{N \times n} } \quad &
    \text{tr} \left( W_o^\top C_{01} \ts W_o \right), \\
    \quad \text{s.t.}  \quad &
    W_o^\top C_{00} W_o = I.
\end{align*}
Defining the Lagrangian $ \mathcal{L}(W_o, \Lambda) = \tr \left(W_o^\top C_{01} W_o \right) - \tr \left( \Lambda \left( W_o^\top C_{00} W_o - I \right) \right)$ and differentiating w.r.t.\ $ W_o $, % using the identity
%     $\frac{\partial}{\partial A} \text{tr}(A^\top B A) = B^\top A + B A \quad \text{and} \quad \frac{\partial}{\partial A} \text{tr}(A^\top B A C) = B^\top A C^\top + B A C$,
see \cite{petersen2008matrix} for matrix derivatives, yields $\nabla_{W_o} \mathcal{L}(W_o, \Lambda) = C_{10} W_o +  C_{01} W_o - 2 \ts C_{00} W_o \Lambda$. Since the operator is self-adjoint, $C_{10} = C_{01}$, and using optimality conditions, we have $C_{01} W_o = C_{00} W_o \Lambda$. Let $\Psi_0 = R(X) \in \mathbb{R}^{N \times m}$ and $\Psi_1 = [(\mathcal{A}R)(x_1), (\mathcal{A}R)(x_2), \dots , (\mathcal{A}R)(x_m)] \in \mathbb{R}^{N \times m}$, then $\widehat{C}_{00} = \Psi_0 \Psi_0^\top$ and $\widehat{C}_{01} = \Psi_0 \Psi_1^\top$. We finally obtain the eigenvalue problem
\begin{equation*}
    \widehat{C}_{01} \ts W_o =  \widehat{C}_{00} W_o \Lambda \implies \widehat{C}_{00}^{+} \ts \widehat{C}_{01} \ts W_o = W_o \Lambda.
\end{equation*}
Hence we have a closed-form solution for the output layer as $W_o$, representing the eigenfunctions of the approximated operator.

\subsection{Non-self-adjoint operators}

We now show how RaNNDy can be used to approximate singular values and functions of a non-self-adjoint operator $\mathcal{A}$. Using the same setting as above for the network with the randomized basis $R(x)$, we use the variational principle for singular values of $\mathcal{A}$. For the data-driven estimation, we start by considering the following ansatz functions for the singular functions $f(x) = W_o^\top R(x)$ and $ g(y) = W_o'^\top R(y)$. Following the derivations in Appendix \ref{appendix:approx_non_self_adjoint}, we get a constrained optimization problem to approximate the singular functions corresponding to the largest $ n $ singular values, i.e.,
\begin{align*}
    \max_{W_o, W_o' \in \mathbb{R}^{N \times n}} & \quad \tr (W_o^\top C_{01} W_o'), \\
    \text{s.t.} & \quad W_o^\top C_{00} W_o = I, \\
    & \quad W_o'^\top C_{11} W_o' = I,
\end{align*}
where the columns of $W_o' \in \mathbb{R}^{N \times n}$ and $W_o \in \mathbb{R}^{N \times n}$ denote the top $n$ left and right singular vectors, respectively. Now, consider the Lagrangian
\begin{align*}
    \mathcal{L}(W_o, W_o', \Lambda, \Lambda') &= \tr(W_o^\top C_{01} W_o') - \tr\left(\frac{1}{2} \Lambda (W_o^\top C_{00} W_o - I) \right) - \tr\left( \frac{1}{2} \Lambda' (W_o'^\top C_{11} W_o' - I) \right).
\end{align*}
% The gradients of $\mathcal{L}$ w.r.t.\ $ W_o $ and $ W_o' $ are
% \begin{align*}
%     \nabla_{W_o} \mathcal{L}(W_o, W_o', \Lambda, \Lambda') &= C_{01} W_o' - C_{00} W_o \Lambda, \\
%     \nabla_{W_o'} \mathcal{L}(W_o, W_o', \Lambda, \Lambda') &= C_{10} W_o - C_{11} W_o' \Lambda'.
% \end{align*}
% Using the optimality conditions, we get
% \begin{align}
%     C_{01} W_o' &= C_{00} W_o \Lambda, \label{eigvalue_fb_1}  \\
%     C_{10} W_o &= C_{11} W_o' \Lambda'. \label{eigvalue_fb_2}
% \end{align}
% Substituting $W_o' = C_{11}^{-1} C_{10} W_o (\Lambda')^{-1}$ from \eqref{eigvalue_fb_2} into \eqref{eigvalue_fb_1} and using $ \Lambda = \Lambda' $,
Solving the above Lagrangian using the optimality conditions, we get
\begin{equation*}
    C_{00}^{+} C_{01} C_{11}^{+} C_{10} W_o = W_o \Lambda^2 \implies  \widehat{C}_{00}^{+} \widehat{C}_{01} \widehat{C}_{11}^{+} \widehat{C}_{10} W_o = W_o \Lambda^2,
\end{equation*}
where $\widehat{C}_{01}, \widehat{C}_{00}, \widehat{C}_{11}$ are the data-driven estimates of the matrices defined in the previous section and $ \widehat{C}_{10} = \widehat{C}_{01}^\top$. Hence, the output layer $W_o$ represents the right singular functions of the operator $\mathcal{A}$, i.e., the eigenfunctions of the operator $\mathcal{A}^*\mathcal{A}$. The above procedure is summarized in Algorithm \ref{alg:eigenfunctions_approximation}. Note that, in order to avoid ill-conditioned matrices, we can, for example, use Tikhonov regularization \cite{golub1999tikhonov}.

\begin{algorithm}
\caption{Eigenfunctions represented by the output layer.}
\begin{algorithmic}
\State \textbf{Initialization:}
\begin{itemize}
 \item Given training data $\big( \psi(x_i), \mathcal{A} \psi(x_i) \big)_{i=1}^m$ for the operator $\mathcal{A}$ to be approximated.
 \item Randomly generate hidden layer weights $W_1, W_2, \dots$, and bias weights, $b_1, b_2, \dots$ from a given distribution to construct a random feature map $R(x)$.
 \item Select an activation function $\sigma$ for the nonlinear transformation.
 \end{itemize}

\State $\blacktriangleright$ Compute $\Psi_0 = [R(x_1), \dots, R(x_m)]$ and $\Psi_1 = [(\mathcal{A}R)(x_1), \dots , (\mathcal{A}R)(x_m)]$.
\State $\blacktriangleright$ Construct the matrices $\widehat{C}_{00} = \Psi_0 \Psi_0^\top$, $\widehat{C}_{11} = \Psi_1 \Psi_1^\top$, $\widehat{C}_{01} = \Psi_0 \Psi_1^\top$ and $\widehat{C}_{10} = \Psi_1 \Psi_0^\top$.
\If{$\mathcal{A}$ is self-adjoint}
    \State $\blacktriangleright$ Solve the eigenvalue problem $ \widehat{C}_{00}^{+} \widehat{C}_{01} W_o = W_o \Lambda $
\Else
    \State $\blacktriangleright$ Solve the eigenvalue problem $ \widehat{C}_{00}^{+} \widehat{C}_{01} \widehat{C}_{11}^{+} \widehat{C}_{10} W_o = W_o \Lambda^2 $
\EndIf

\State $\blacktriangleright$ Sort the eigenvectors in $W_o$ according to the eigenvalues and select the top $n$ eigenvectors to get the approximation of the dominant $n$ eigenfunctions.
\end{algorithmic}
\label{alg:eigenfunctions_approximation}
\end{algorithm}

\section{Numerical experiments}
\label{sec:numerical_experiments}

We now apply the proposed techniques to different operators and various benchmark problems and also compare RaNNDy and VAMPnets in terms of the computational costs.

\subsection{Overdamped Langevin SDE}

\begin{figure}
    \centering
    \begin{minipage}[t]{0.32\textwidth}
        \subfloat[][\label{subfig:ou_eigvals}]{\includegraphics[width=0.9\textwidth]{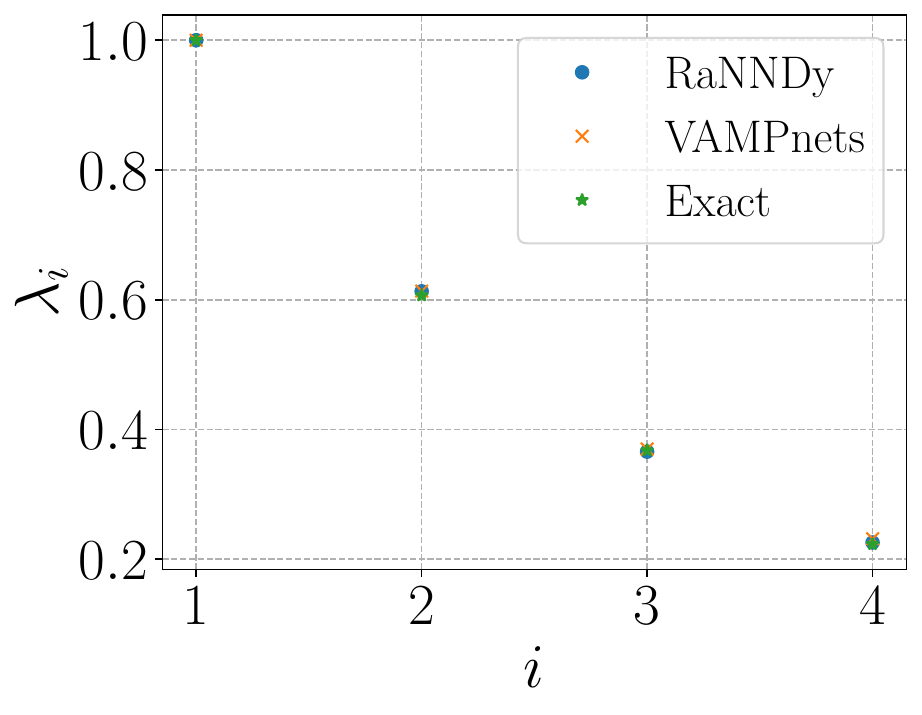}
    }
    \end{minipage}
    \begin{minipage}[t]{0.32\textwidth}
        \subfloat[][\label{subfig:ou_eigfuncs}]{\includegraphics[width=0.9\textwidth]{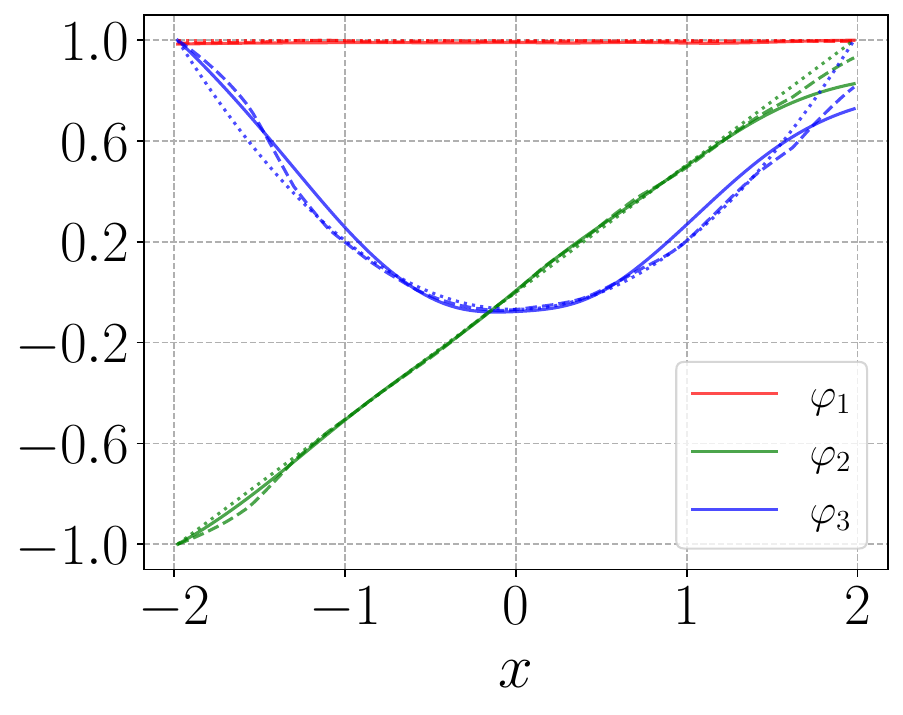}}
    \end{minipage}
    \begin{minipage}[t]{0.32\textwidth}
        \subfloat[][\label{subfig:ou_error_comp}]{\includegraphics[width=0.9\textwidth]{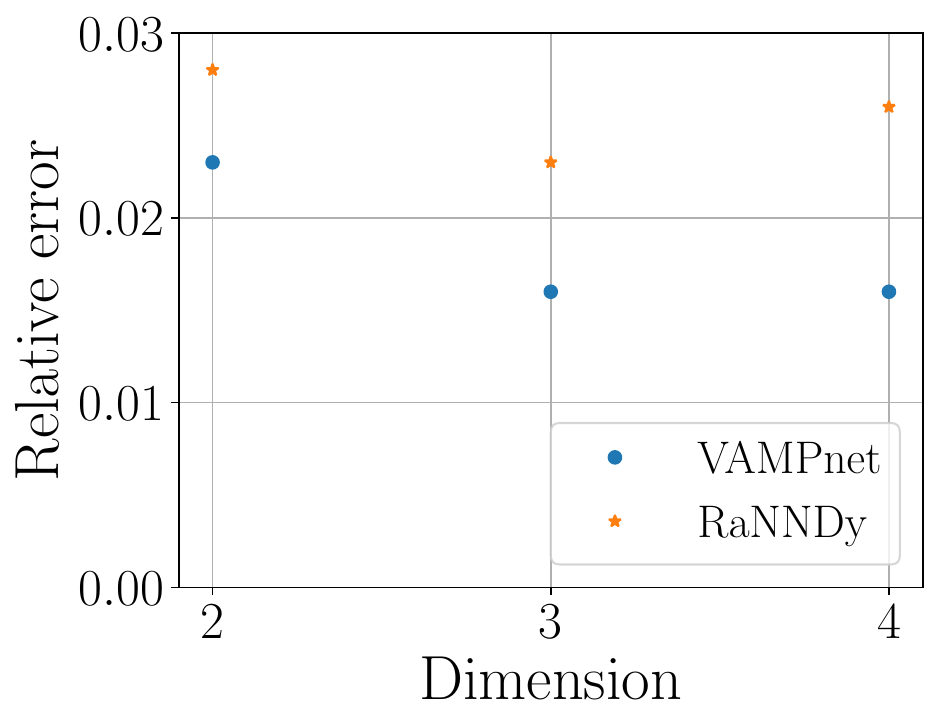}}
    \end{minipage}
    % \begin{minipage}[t]{0.24\textwidth}
    %     \subfloat[][\label{subfig:grid_serach_ou1d}]{\includegraphics[width=0.9\textwidth]{pics/grid_search_ou1d.pdf}}
    % \end{minipage}
    \caption{OU process results. (a) Four dominant eigenvalues showing a perfect match between RaNNDy, VAMPnets, and the exact eigenvalues. (b)~Three dominant eigenfuntions. The solid lines represent RaNNDy approximations, dashed lines are VAMPnets, and dotted lines are the exact eigenfunctions. (c)~Comparison of approximation errors for multi-dimensional OU processes.}
    \label{fig:ou_results}
\end{figure}

For the one-dimensional Ornstein--Uhlenbeck (OU) process, we can compute eigenvalues and eigenfunctions of the Koopman operator analytically. The details on the OU process, governing SDE, and the experimental setup are provided in Appendix \ref{appendix:overdamped_sde}. Using the same setup, the numerical approximations computed by RaNNDy and VAMPnets for the eigenvalues and eigenfunctions are shown in Figures \ref{subfig:ou_eigvals} and \ref{subfig:ou_eigfuncs}, respectively, matching the analytical solutions. Since we create the data using a single long trajectory containing very few data points near the boundaries of the domain, the approximations are less accurate in these regions. Furthermore, for multi-dimensional OU processes, see~\cite{zhang2021computing, tabish_thesis_2026}, we compute the analytical eigenvalues and compare them with their numerically computed counterparts, see Figure~\ref{subfig:ou_error_comp}. We can use a grid-search strategy to select optimal distributions for the initialization of the weights and biases of the randomized neural network using the variational score as a metric.

\paragraph{Further analysis:} In Appendix \ref{appendix:overdamped_sde}, we provide a more detailed comparison between RaNNDy and VAMPnets regarding the accuracy and run times with different settings. These results support our claim that with the right initialization of the network, we can skip deep learning and avoid problems such as slow convergence, sensitivity to hyperparameters, loss of interpretability of basis functions, etc. Moreover, we show the approximation of singular value decomposition using RaNNDy for the case when the associated transfer operators are not self-adjoint using the Bickley jet system in Appendix \ref{appendix:bickley_jet}.

\subsection{Protein folding problems}

We now show how RaNNDy can be applied to high-dimensional protein folding processes. We are in particular interested in identifying the folded and unfolded structures of molecules, see~\cite{Schuette_Klus_Hartmann_2023} for a detailed introduction to molecular dynamics. The goal is to gain a better understanding of biological processes, which is of significant importance in the development of new drugs or treatments \cite{dill2008protein}. We apply the Koopman analysis using the proposed framework to different data sets acquired from \href{https://www.deshawresearch.com/}{\color{blue}{D.E. Shaw Research}}, see \cite{lindorff2011fast}. We apply RaNNDy, described in Algorithm \ref{alg:eigenfunctions_approximation}, and VAMPnets with $100$ epochs of the Adam optimizer \cite{kingma2014adam} to train the network. In order to ensure the comparison is fair, we select the same network architecture consisting of three dense hidden layers comprising 256, 512, and 256 neurons and approximate the first few dominant eigenvalues and eigenfunctions of the Koopman operator.

\paragraph{Chignolin (CLN025): }

\begin{figure}
    \centering
    \begin{minipage}[t]{0.45\textwidth}
        \subfloat[][\label{subfig:chig_folded_states}]{\includegraphics[width=0.55\textwidth]{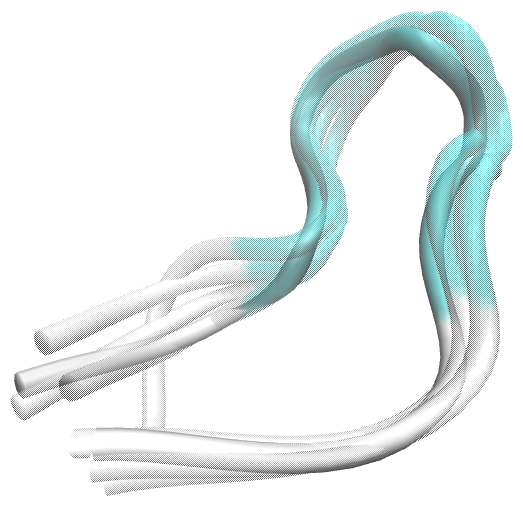}}
    \end{minipage}
    \begin{minipage}[t]{0.45\textwidth}
        \subfloat[][\label{subfig:chig_unfolded_states}]{\includegraphics[width=0.7\textwidth]{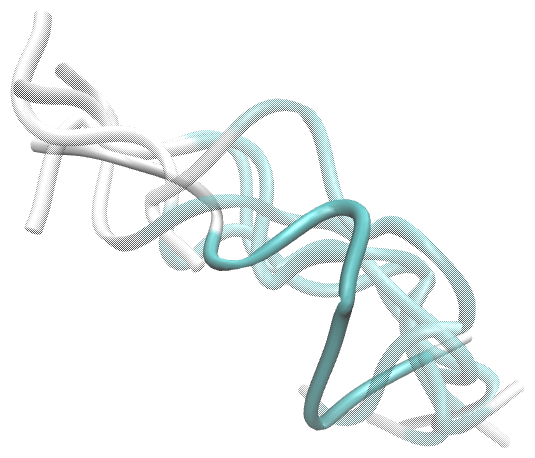}}
    \end{minipage} \\
    \begin{minipage}[t]{0.45\textwidth}
        \subfloat[][\label{subfig:chig_folded_contact}]{\includegraphics[width=0.9\textwidth]{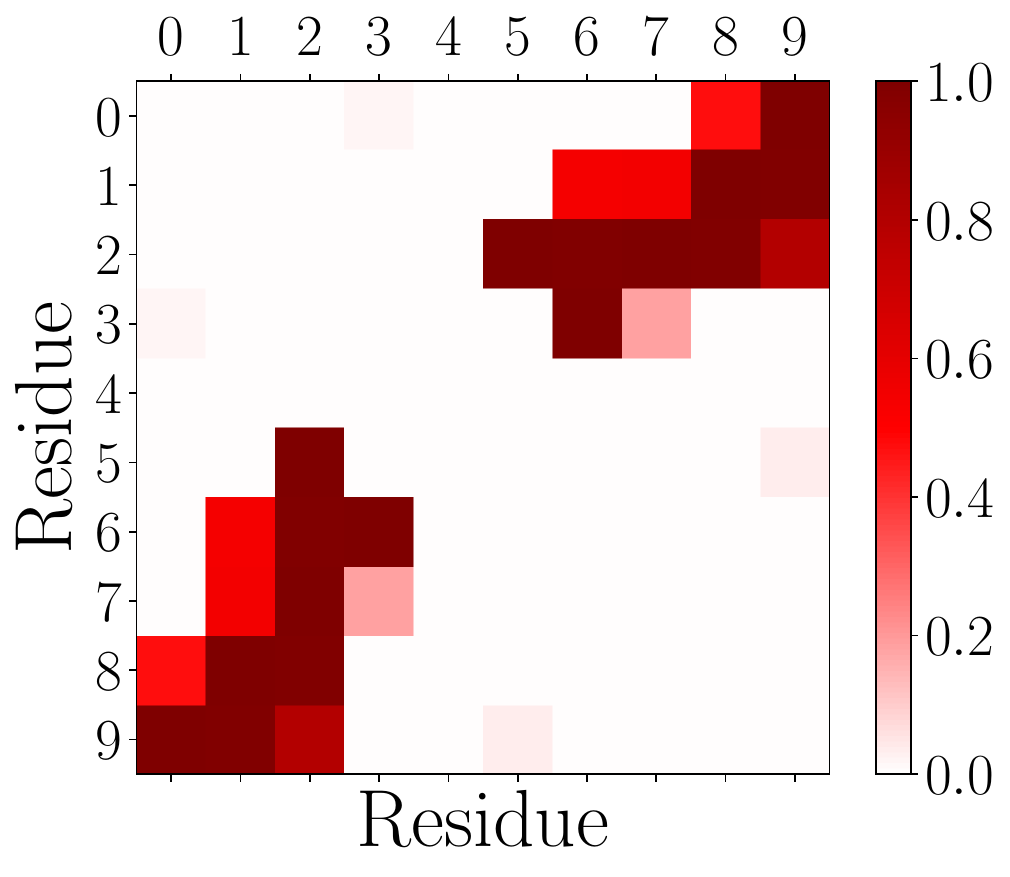}}
    \end{minipage}
    \begin{minipage}[t]{0.45\textwidth}
        \subfloat[][\label{subfig:chig_unfolded_contact}]{\includegraphics[width=0.9\textwidth]{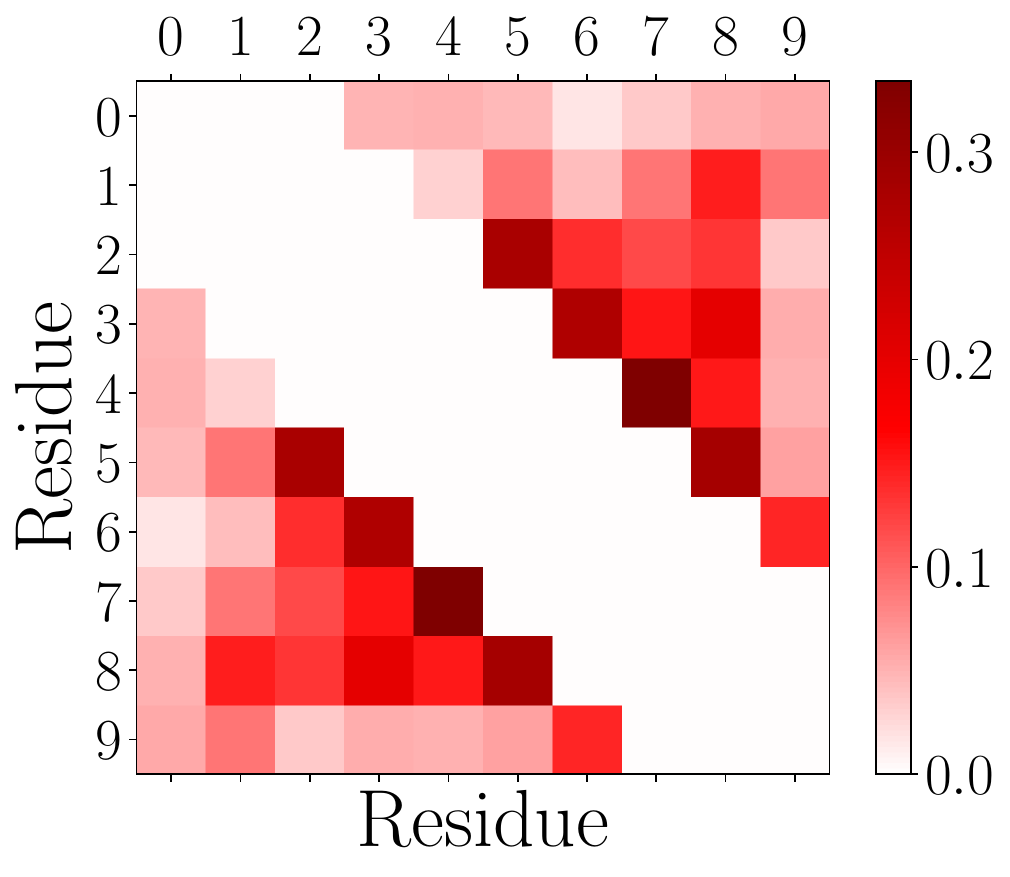}}
    \end{minipage}
    \caption{RaNNDy results for Chignolin. (a) \& (b) Folded and unfolded states. (c) \& (d) Frequency of contacts between different residue pairs over all the identified folded and unfolded states, respectively.}
    \label{fig:chig_results}
\end{figure}

Chignolin is a mini-protein molecule consisting of $10$ residues and $166$ atoms. The trajectory we consider is $1.06 \cdot 10^8$\ts ps long. We subsample the trajectory to create the training data $\{x_i, y_i\}_{i=1}^m$ using the lag time $\tau=9911.5$\ts ps, i.e., the gap between $x_i$ and $y_i$ is $50$ frames. The subsampled trajectory data is then transformed using contact maps, which measure the distances between different pairs of residues, resulting in $X, Y \in \R^{28 \times 10 \ts 694}$. A few folded and unfolded states, identified by RaNNDy, are shown in Figures \ref{subfig:chig_folded_states} and \ref{subfig:chig_unfolded_states}, respectively. Figures \ref{subfig:chig_folded_contact} and \ref{subfig:chig_unfolded_contact} show the frequency of contacts between different residues of the molecule for the sets of identified folded and unfolded states.

\paragraph{Protein NuG2: }

\begin{figure}
    \centering
    \begin{minipage}[t]{0.42\textwidth}
        \subfloat[][\label{subfig:nug2_folded_states}]{\includegraphics[width=0.9\textwidth]{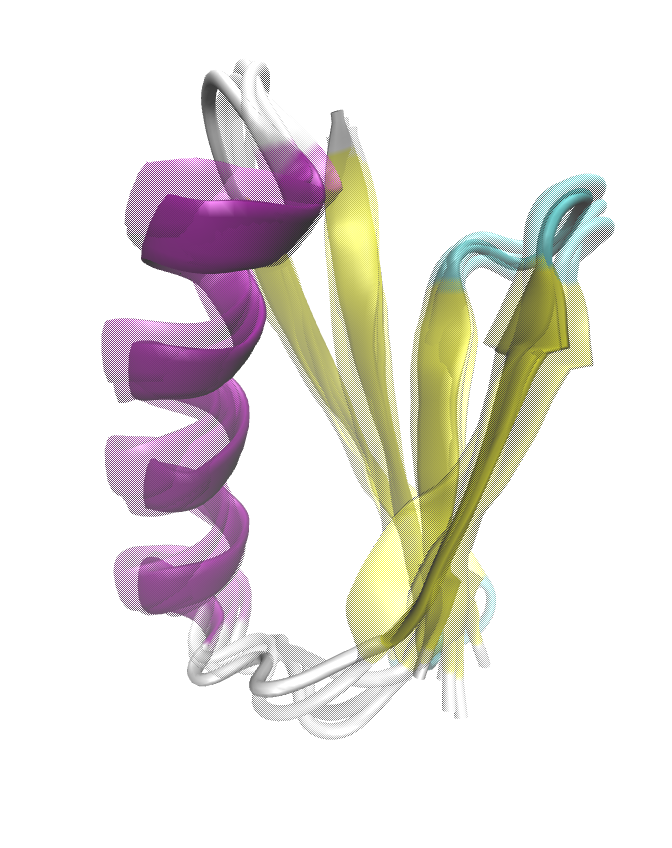}}
    \end{minipage}
    \begin{minipage}[t]{0.42\textwidth}
        \subfloat[][\label{subfig:nug2_unfolded_states}]{\includegraphics[width=0.8\textwidth]{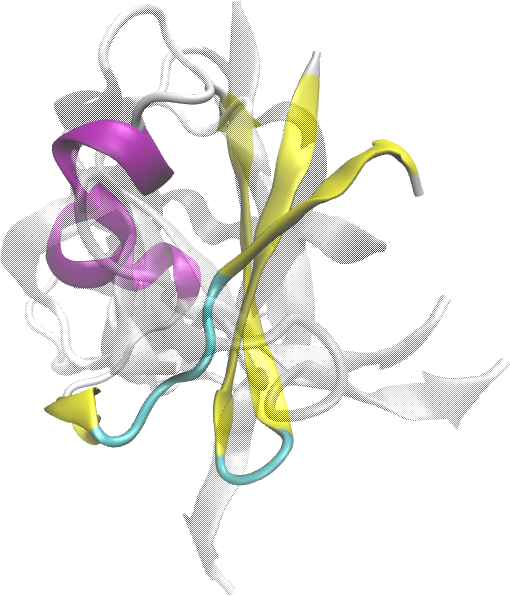}}
    \end{minipage} \\ %\\[1ex]
    \begin{minipage}[t]{0.45\textwidth}
        \subfloat[][\label{subfig:nug2_folded_contact}]{\\[1ex]\includegraphics[width=0.9\textwidth]{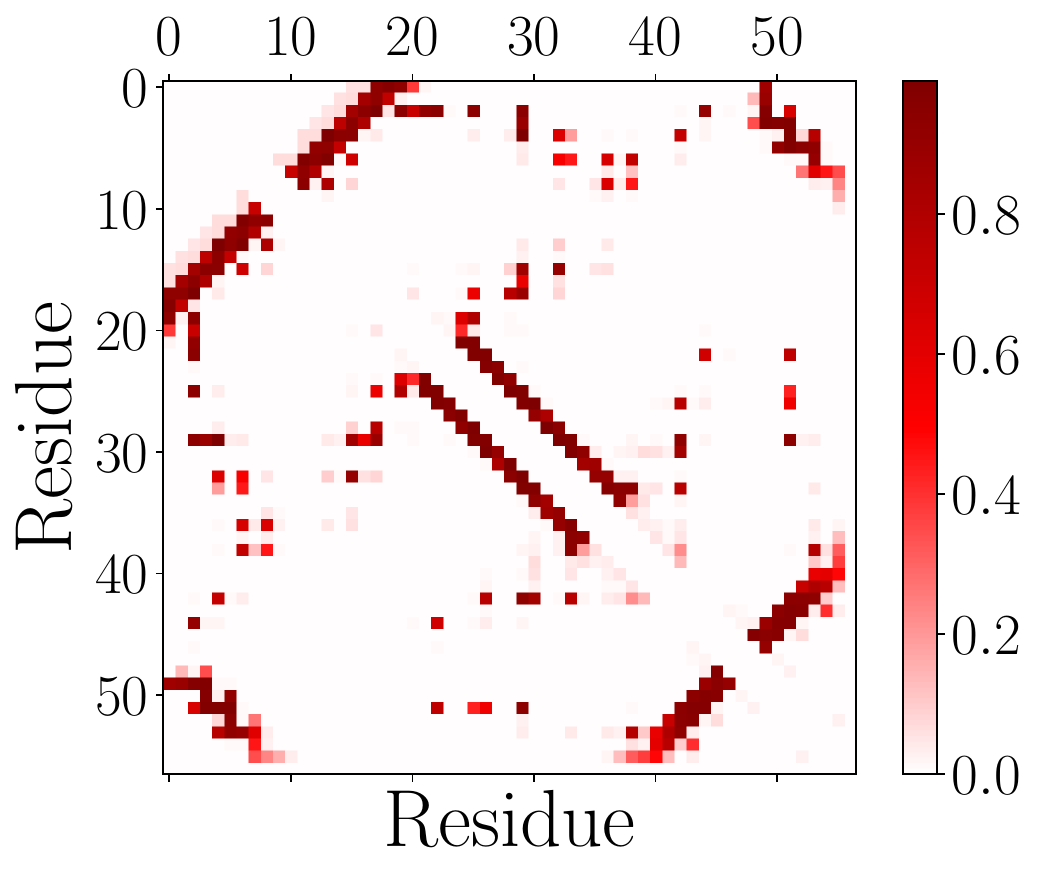}}
    \end{minipage}
    \begin{minipage}[t]{0.45\textwidth}
        \subfloat[][\label{subfig:nug2_unfolded_contact}]{\\[1ex]\includegraphics[width=0.9\textwidth]{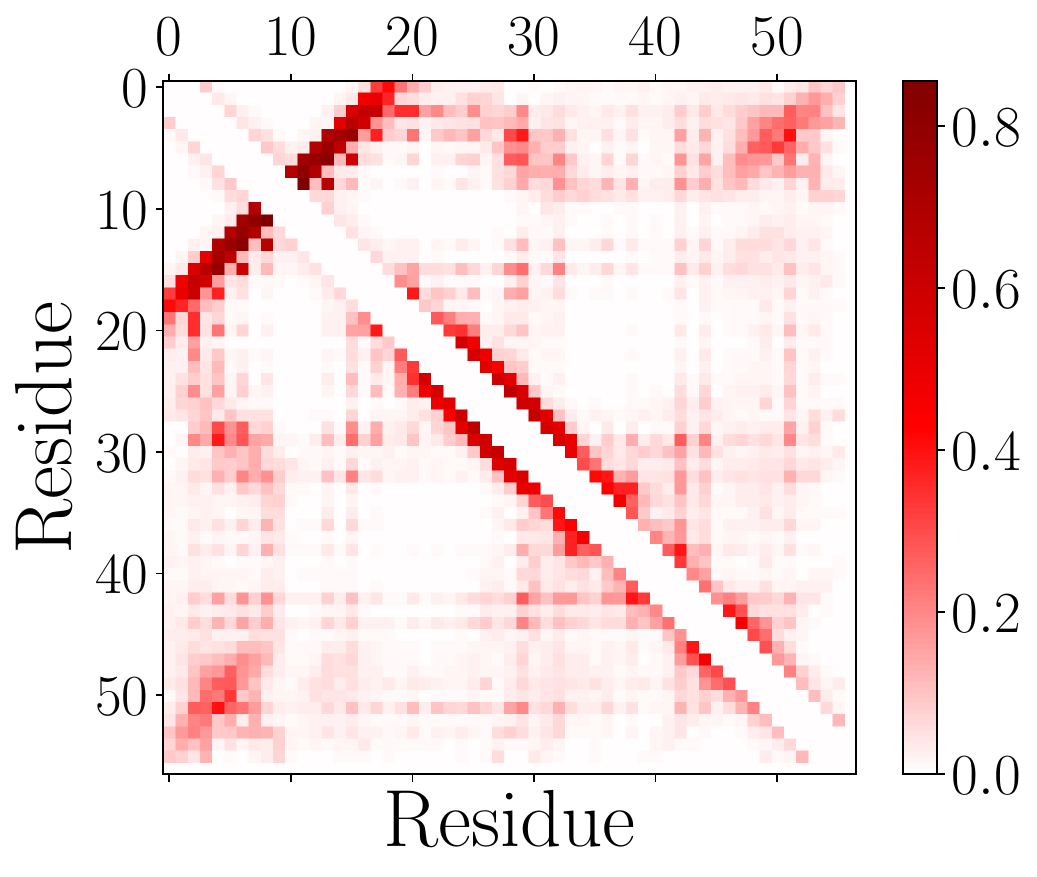}}
    \end{minipage}
    \caption{RaNNDy results for NuG2. (a) \& (b) Some folded and unfolded states of the NuG2 protein. (c) \& (d) Frequency of contacts between different residue pairs over all the identified folded and unfolded states, respectively.}
    \label{fig:nug2_results}
\end{figure}

Protein G (NuG2) consists of $56$ residues. The total trajectory is $3.680002 \cdot 10^8$\ts ps long. We consider a subset of the simulation trajectories and subsample the selected $929251$ frames to create the training data $\{x_i, y_i\}_{i=1}^m$. The lag time is a gap of $50$ frames between each $x_i$ and $y_i$. We again compute the contact maps to get the data matrices $X, Y \in \R^{ 1431 \times 18 \ts 585}$. That is, the dynamics are really high-dimensional in this case. Figure \ref{fig:nug2_results} presents the folded and unfolded states and the contact frequencies of the NuG2 protein molecule identified using RaNNDy. Additionally, we provide the videos of the molecule's folded and unfolded states identified using RaNNDy in the supplementary material.

\begin{figure}
    \centering
        \subfloat[][\label{subfig:chig_eigfuncs}]{\includegraphics[width=0.31\textwidth]{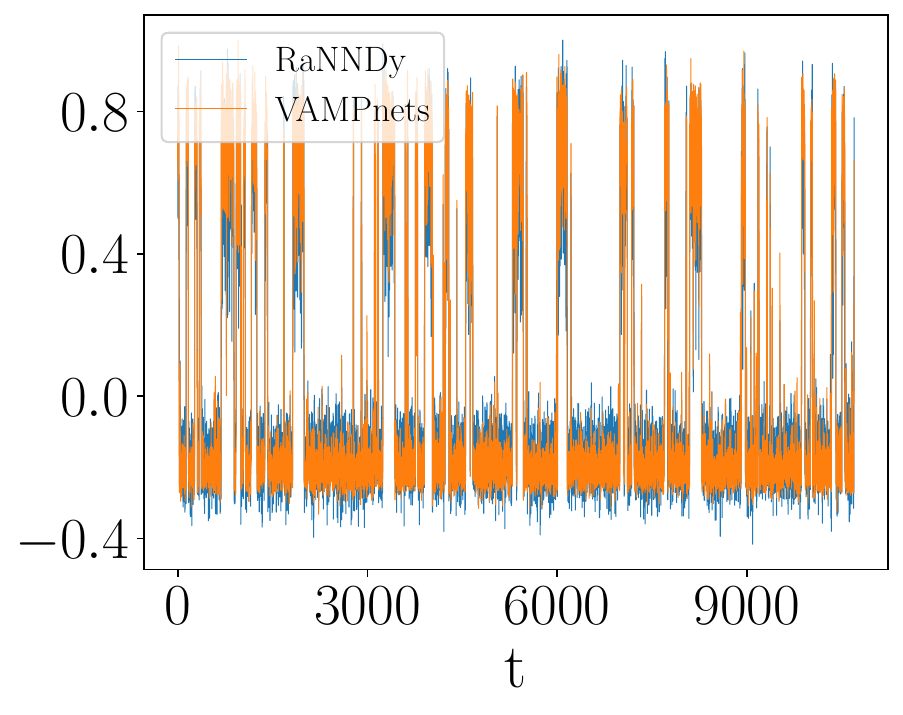}
    }
        \subfloat[][\label{subfig:nug2_eigfuncs}]{\includegraphics[width=0.31\textwidth]{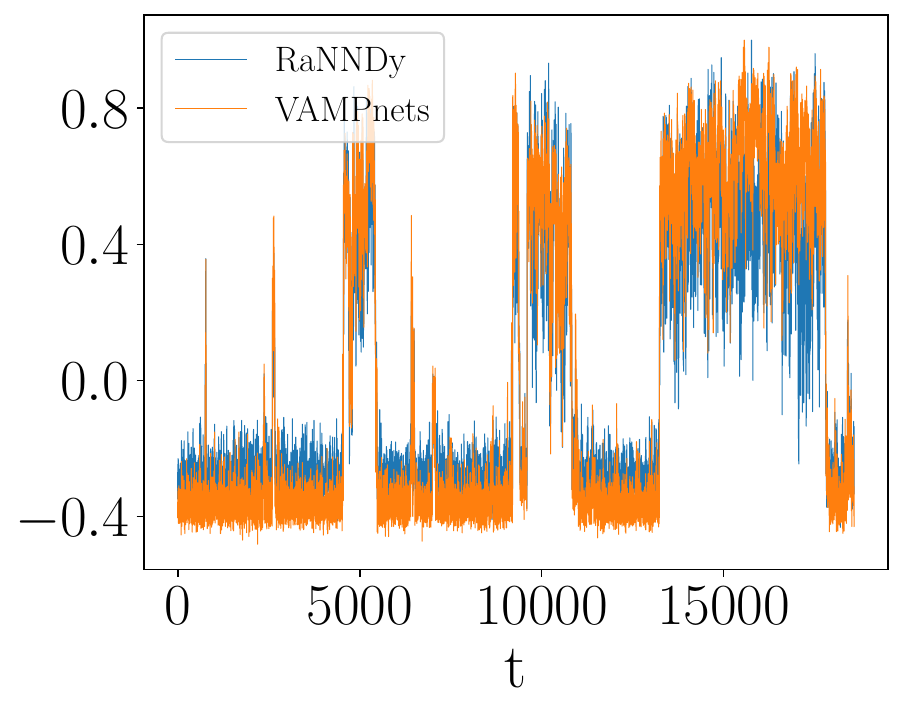}}
        \subfloat[][\label{subfig:nug2_uncertainty_trend}]{\includegraphics[width=0.3\textwidth]{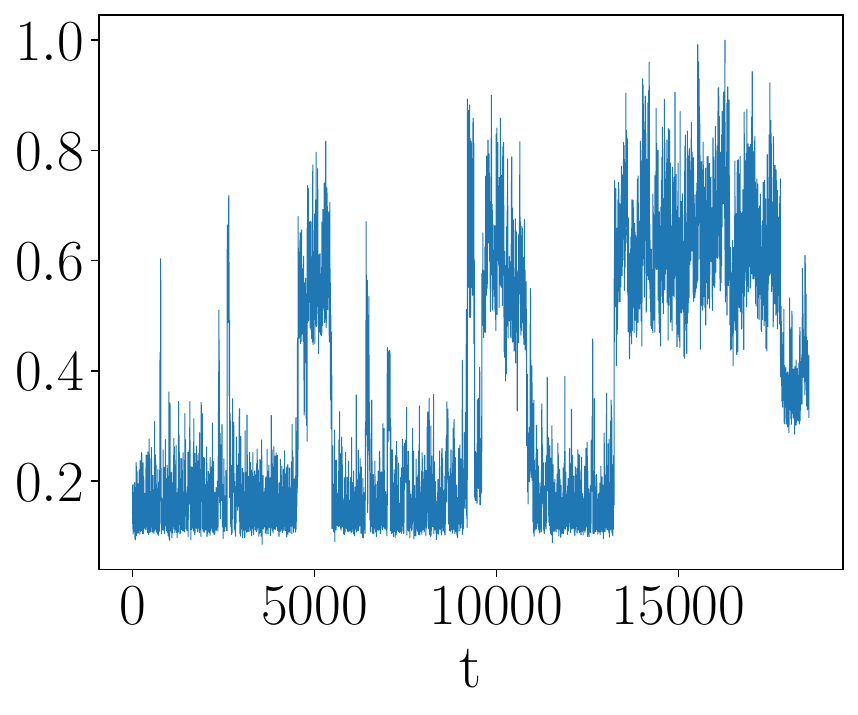}}
    \caption{Protein folding results. Second dominant eigenfunction evaluated in the trajectory data using RaNNDy and VAMPnets for (a) the Chignolin mini-protein and (b) the NuG2 protein. (c)~Uncertainty associated with the second eigenfunction of the NuG2 protein, indicating a higher uncertainty for the unfolded states. Unfolded states are more chaotic (as can be seen from the eigenfunction and also Figure~\ref{subfig:nug2_unfolded_states}). Hence, we see a high uncertainty in that region.}
    \label{fig:md_eigfuncs}
\end{figure}

\paragraph{Comparison: }
Figure~\ref{fig:md_eigfuncs} shows the second eigenfunctions associated with the two protein molecules above identified using RaNNDy and VAMPnets (with the same network architecture). We can see that the behavior of the eigenfunctions computed by both methods is very similar. Furthermore, in Table \ref{tab:protein_folding_results}, we compare RaNNDy and VAMPnets in terms of the computational costs and accuracy in identifying the two states of the molecules. The contact maps of the two states of the above molecules identified using RaNNDy again highlight its accuracy.
\begin{table}
    \centering
    \caption{RaNNDy and VAMPnets run times for protein-folding problems.}
    \resizebox{\textwidth}{!}{
    \begin{tabular}{c c c c c c c c c}
        \toprule
         & \multicolumn{2}{c}{Computational time (in sec.)} & \multicolumn{2}{c}{Identified folded frames} & \multicolumn{2}{c}{Identified unfolded frames} \\
        \cmidrule(lr){2-3} \cmidrule(lr){4-5} \cmidrule(lr){6-7}
        Protein & RaNNDy & VAMPnets & RaNNDy & VAMPnets & RaNNDy & VAMPnets \\
        \midrule
        Chignolin & 0.52 & 60.44 & 8282 & 8296 & 2412 & 2398  \\
        NuG2 & 1.34 & 139.98 & 11706 & 11728 & 6879 & 6857 \\
        \bottomrule
    \end{tabular}
    }
    \label{tab:protein_folding_results}
\end{table}

% \begin{figure}
% \centering

% \begin{minipage}{0.06\textwidth}
%     \centering
%     \rotatebox{90}{\Large VAMP score}
% \end{minipage}
% \hfill
% \begin{minipage}{0.9\textwidth}
%     \centering
%     \subfloat[\label{subfig:ou2dcomp_normal1}]
%     {\includegraphics[width=0.2\linewidth]{pics/ou2dcomp_normal.pdf}}
%     \subfloat[\label{subfig:ou2dcomp_uniform1}]
%     {\includegraphics[width=0.2\linewidth]{pics/ou2dcomp_uniform.pdf}}
%     \subfloat[\label{subfig:ou2dcomp_lecun_normal1}]
%     {\includegraphics[width=0.2\linewidth]{pics/ou2dcomp_lecun_normal.pdf}}

%     \subfloat[\label{subfig:ou2dcomp_normal_poor1}]
%     {\includegraphics[width=0.2\linewidth]{pics/ou2dcomp_normal_poor.pdf}}

%     {\Large Epochs (time in sec.)}

% \end{minipage}
% \caption{RaNNDy and VAMPnet comparison for approximation of the sum of five dominant eigenvalues. The weights in VAMPnets are initialized from (a) normal distribution. (b) uniform distribution. (c) LeCun normal distribution. (d) Divergence of VAMPnets. }
% \label{fig:multidim_ou1}

% \end{figure}

\section{Conclusion}
\label{sec:conclusion}

We proposed a novel data-driven framework, called RaNNDy, for the approximation of linear operators and their eigenvalues and eigenfunctions or singular values and singular functions. Built on randomized neural networks, RaNNDy generates randomized features through the hidden layers of the network, which are kept fixed. Only the output layer is trained to approximate the operator. We introduced loss functions to train RaNNDy using variational principles. Furthermore, we derived a closed-form solution for the output layer of RaNNDy that represents the eigenfunctions of the operator. This novel approach has the advantage that it significantly reduces the training time and computational costs, making the network simple and easy to implement, while mitigating the risk of common problems arising in the training of deep neural networks. The proposed framework furthermore enables us to use ensemble learning techniques to quantify uncertainties in the operator approximation. With a proper initialization of weights, we show that the numerical results of the proposed approach are efficient and comparable to state-of-the-art deep learning methods across different systems. The results imply that the randomly generated features are suitable basis functions and that, by using a grid-search type method for selecting proper distributions for initial weights, similar accuracy can be achieved. This also makes the algorithm more interpretable, as we know the structure of basis functions, i.e., the exact distribution of weights and biases that parametrize the basis.

\paragraph{Limitations and future work:} RaNNDy can be viewed as a compromise between data-driven models that utilize fixed basis functions such as EDMD and fully optimizable neural networks such as VAMPnets. Although VAMPnets might in general be more flexible and have better approximation properties, RaNNDy is computationally efficient and provides comparable accuracy. There are, however, some open problems. The first and most important is to select the distribution of weights and biases for the randomized part of the network. Choosing distributions that make the matrix representations of the operators ill-conditioned will negatively affect the approximation. This analysis could be extremely important in making RaNNDy more robust and generalizable \cite{arpit_2019, mishkin2015all, datta2020survey}, similar to recent work on efficient and data-informed sampling of hidden layers' weights for deep and shallow networks \cite{bolager_2023}. We present detailed ablation studies in Appendix \ref{appendix:ablation_studies}, comparing different initializations of RaNNDy and illustrating that a grid search for finding the optimal distributions of weights still takes significantly less time than training the whole network. For a specific choice of activation function and distribution of weights, random Fourier feature models \cite{rahimi2007random, nuske2023efficient} can be considered as a special case of RaNNDy. Hence, an interesting future research direction could be to explore the relationships of different activation functions and  weight distributions with kernel-based techniques and random Fourier features. The code can be found at \href{https://github.com/tabishhub/ranndy}{\color{blue}https://github.com/tabishhub/ranndy}.

\section*{Acknowledgments}
MT was supported by the EPSRC Centre for Doctoral Training in Mathematical Modeling, Analysis and Computation (MAC-MIGS) funded by the UK Engineering and Physical Sciences Research Council (grant EP/S023291/1), Heriot--Watt University and the University of Edinburgh. We would like to thank D.E.\ Shaw Research for providing the protein molecule data.

\bibliographystyle{unsrtnat}
\bibliography{references}

\begin{thebibliography}{67}
\providecommand{\natexlab}[1]{#1}
\providecommand{\url}[1]{\texttt{#1}}
\expandafter\ifx\csname urlstyle\endcsname\relax
  \providecommand{\doi}[1]{doi: #1}\else
  \providecommand{\doi}{doi: \begingroup \urlstyle{rm}\Url}\fi

\bibitem[Klus and Conrad(2024)]{klus2024dynamical}
S.~Klus and N.D. Conrad.
\newblock Dynamical systems and complex networks: A {K}oopman operator
  perspective.
\newblock \emph{Journal of Physics: Complexity}, 5\penalty0 (4):\penalty0
  041001, 2024.
\newblock \doi{10.1088/2632-072X/ad9e60}.

\bibitem[Dellnitz and Junge(1999)]{dellnitz1999approximation}
M.~Dellnitz and O.~Junge.
\newblock On the approximation of complicated dynamical behavior.
\newblock \emph{SIAM Journal on Numerical Analysis}, 36\penalty0 (2):\penalty0
  491--515, 1999.
\newblock \doi{10.1137/S0036142996313002}.

\bibitem[Mezi{\'c}(2005)]{mezic2005spectral}
I.~Mezi{\'c}.
\newblock Spectral properties of dynamical systems, model reduction and
  decompositions.
\newblock \emph{Nonlinear Dynamics}, 41:\penalty0 309--325, 2005.
\newblock \doi{10.1007/s11071-005-2824-x}.

\bibitem[Lasota and Mackey(2013)]{lasota2013chaos}
A.~Lasota and M.C. Mackey.
\newblock \emph{Chaos, fractals, and noise: {S}tochastic aspects of dynamics},
  volume~97.
\newblock Springer New York, NY, 2013.
\newblock \doi{10.1007/978-1-4612-4286-4}.

\bibitem[Klus et~al.(2016)Klus, Koltai, and Sch{\"u}tte]{KKS16}
S.~Klus, P.~Koltai, and C.~Sch{\"u}tte.
\newblock On the numerical approximation of the {P}erron--{F}robenius and
  {K}oopman operator.
\newblock \emph{Journal of Computational Dynamics}, 3\penalty0 (1):\penalty0
  51--79, 2016.
\newblock \doi{10.3934/jcd.2016003}.

\bibitem[Ulam(1960)]{Ulam60}
S.~M. Ulam.
\newblock \emph{A Collection of Mathematical Problems}.
\newblock Interscience Publisher NY, 1960.

\bibitem[Williams et~al.(2015{\natexlab{a}})Williams, Kevrekidis, and
  Rowley]{williams2015data}
M.O. Williams, I.G. Kevrekidis, and C.W. Rowley.
\newblock A data-driven approximation of the {K}oopman operator: Extending
  dynamic mode decomposition.
\newblock \emph{Journal of Nonlinear Science}, 25:\penalty0 1307--1346,
  2015{\natexlab{a}}.
\newblock \doi{10.1007/s00332-015-9258-5}.

\bibitem[Tabish et~al.(2025)Tabish, Chada, and Klus]{TABISH2025134822}
M.~Tabish, N.K. Chada, and S.~Klus.
\newblock Learning dynamical systems from data: {G}radient-based dictionary
  optimization.
\newblock \emph{Physica D: Nonlinear Phenomena}, 481:\penalty0 134822, 2025.
\newblock \doi{10.1016/j.physd.2025.134822}.

\bibitem[Williams et~al.(2015{\natexlab{b}})Williams, Rowley, and
  Kevrekidis]{WRK15}
M.~O. Williams, C.~W. Rowley, and I.~G. Kevrekidis.
\newblock A kernel-based method for data-driven {K}oopman spectral analysis.
\newblock \emph{Journal of Computational Dynamics}, 2\penalty0 (2):\penalty0
  247--265, 2015{\natexlab{b}}.
\newblock \doi{10.3934/jcd.2015005}.

\bibitem[Klus et~al.(2020{\natexlab{a}})Klus, Schuster, and Muandet]{KSM20}
S.~Klus, I.~Schuster, and K.~Muandet.
\newblock Eigendecompositions of transfer operators in reproducing kernel
  {H}ilbert spaces.
\newblock \emph{Journal of Nonlinear Science}, 2020{\natexlab{a}}.
\newblock \doi{10.1007/s00332-019-09574-z}.

\bibitem[Klus et~al.(2020{\natexlab{b}})Klus, N\"uske, Peitz, Niemann,
  Clementi, and Sch\"utte]{KNPNCS20}
S.~Klus, F.~N\"uske, S.~Peitz, J.-H. Niemann, C.~Clementi, and C.~Sch\"utte.
\newblock Data-driven approximation of the {K}oopman generator: {M}odel
  reduction, system identification, and control.
\newblock \emph{Physica D: Nonlinear Phenomena}, 406:\penalty0 132416,
  2020{\natexlab{b}}.
\newblock \doi{10.1016/j.physd.2020.132416}.

\bibitem[Li et~al.(2017)Li, Dietrich, Bollt, and Kevrekidis]{li2017extended}
Q.~Li, F.~Dietrich, E.~M. Bollt, and I.~G. Kevrekidis.
\newblock Extended dynamic mode decomposition with dictionary learning: A
  data-driven adaptive spectral decomposition of the {K}oopman operator.
\newblock \emph{Chaos: An Interdisciplinary Journal of Nonlinear Science},
  27\penalty0 (10), 2017.
\newblock \doi{10.1063/1.4993854}.

\bibitem[Yeung et~al.(2019)Yeung, Kundu, and Hodas]{enoch2019}
E.~Yeung, S.~Kundu, and N.~Hodas.
\newblock Learning deep neural network representations for {K}oopman operators
  of nonlinear dynamical systems.
\newblock In \emph{2019 American Control Conference (ACC)}, pages 4832--4839,
  2019.
\newblock \doi{10.23919/ACC.2019.8815339}.

\bibitem[Gulina and Mauroy(2021)]{gulina2021two}
M.~Gulina and A.~Mauroy.
\newblock Two methods to approximate the {K}oopman operator with a reservoir
  computer.
\newblock \emph{Chaos: An Interdisciplinary Journal of Nonlinear Science},
  31\penalty0 (2), 2021.
\newblock \doi{10.1063/5.0026380}.

\bibitem[Xiao et~al.(2022)Xiao, Zhang, Xu, Liu, and Liu]{xiao2022deep}
Y.~Xiao, X.~Zhang, X.~Xu, X.~Liu, and J.~Liu.
\newblock Deep neural networks with {K}oopman operators for modeling and
  control of autonomous vehicles.
\newblock \emph{IEEE Transactions on Intelligent Vehicles}, 8\penalty0
  (1):\penalty0 135--146, 2022.

\bibitem[Mardt et~al.(2018)Mardt, Pasquali, Wu, and No{\'e}]{mardt2018vampnets}
A.~Mardt, L.~Pasquali, H.~Wu, and F.~No{\'e}.
\newblock {VAMP}nets for deep learning of molecular kinetics.
\newblock \emph{Nature communications}, 9\penalty0 (1):\penalty0 5, 2018.
\newblock \doi{10.1038/s41467-017-02388-1}.

\bibitem[Wu and No{\'e}(2020)]{wu2020variational}
H.~Wu and F.~No{\'e}.
\newblock Variational approach for learning {M}arkov processes from time series
  data.
\newblock \emph{Journal of Nonlinear Science}, 30\penalty0 (1):\penalty0
  23--66, 2020.
\newblock \doi{10.1007/s00332-019-09567-y}.

\bibitem[No{\'e} and Nüske(2013)]{noe2013variational}
F.~No{\'e} and F.~Nüske.
\newblock A variational approach to modeling slow processes in stochastic
  dynamical systems.
\newblock \emph{Multiscale Modeling \& Simulation}, 11\penalty0 (2):\penalty0
  635--655, 2013.
\newblock \doi{10.1137/110858616}.

\bibitem[Gori et~al.(1992)Gori, Tesi, et~al.]{gori1992problem}
M.~Gori, A.~Tesi, et~al.
\newblock On the problem of local minima in backpropagation.
\newblock \emph{IEEE Transactions on Pattern Analysis and Machine
  Intelligence}, 14\penalty0 (1):\penalty0 76--86, 1992.
\newblock \doi{10.1109/34.107014}.

\bibitem[{Te Braake} and {Van Straten}(1995)]{TEBRAAKE199571}
H.A.B. {Te Braake} and G.~{Van Straten}.
\newblock Random activation weight neural net ({RAWN}) for fast non-iterative
  training.
\newblock \emph{Engineering Applications of Artificial Intelligence},
  8\penalty0 (1):\penalty0 71--80, 1995.
\newblock \doi{10.1016/0952-1976(94)00056-S}.

\bibitem[Staib et~al.(2019)Staib, Reddi, Kale, Kumar, and
  Sra]{staib2019escaping}
M.~Staib, S.~Reddi, S.~Kale, S.~Kumar, and S.~Sra.
\newblock Escaping saddle points with adaptive gradient methods.
\newblock In \emph{International Conference on Machine Learning}, pages
  5956--5965. PMLR, 2019.

\bibitem[Kingma and Ba(2015)]{kingma2014adam}
Diederik~P. Kingma and Jimmy Ba.
\newblock Adam: A method for stochastic optimization.
\newblock In \emph{International Conference on Learning Representations
  (ICLR)}, 2015.
\newblock URL \url{https://arxiv.org/abs/1412.6980}.

\bibitem[Pascanu et~al.(2013)Pascanu, Mikolov, and
  Bengio]{pascanu2013difficulty}
R.~Pascanu, T.~Mikolov, and Y.~Bengio.
\newblock On the difficulty of training recurrent neural networks.
\newblock In \emph{International Conference on Machine Learning}, pages
  1310--1318. PMLR, 2013.

\bibitem[Zhang and Suganthan(2016{\natexlab{a}})]{zhang2016survey}
L.~Zhang and P.N. Suganthan.
\newblock A survey of randomized algorithms for training neural networks.
\newblock \emph{Information Sciences}, 364:\penalty0 146--155,
  2016{\natexlab{a}}.
\newblock \doi{10.1016/j.ins.2016.01.039}.

\bibitem[Suganthan and Katuwal(2021)]{suganthan2021origins}
P.N. Suganthan and R.~Katuwal.
\newblock On the origins of randomization-based feedforward neural networks.
\newblock \emph{Applied Soft Computing}, 105:\penalty0 107239, 2021.
\newblock \doi{10.1016/j.asoc.2021.107239}.

\bibitem[Malik et~al.(2023)Malik, Gao, Ganaie, Tanveer, and
  Suganthan]{malik2023random}
A.K. Malik, R.~Gao, M.A. Ganaie, M.~Tanveer, and P.N. Suganthan.
\newblock Random vector functional link network: recent developments,
  applications, and future directions.
\newblock \emph{Applied Soft Computing}, 143:\penalty0 110377, 2023.
\newblock \doi{10.1016/j.asoc.2023.110377}.

\bibitem[Weipeng et~al.(2018)Weipeng, Xizhao, Zhong, and Jinzhu]{CAO2018278}
C.~Weipeng, W~Xizhao, M.~Zhong, and G.~Jinzhu.
\newblock A review on neural networks with random weights.
\newblock \emph{Neurocomputing}, 275:\penalty0 278--287, 2018.
\newblock ISSN 0925-2312.
\newblock \doi{https://doi.org/10.1016/j.neucom.2017.08.040}.

\bibitem[Zhang and Suganthan(2016{\natexlab{b}})]{ZHANG2016146}
L.~Zhang and P.N. Suganthan.
\newblock A survey of randomized algorithms for training neural networks.
\newblock \emph{Information Sciences}, 364-365:\penalty0 146--155,
  2016{\natexlab{b}}.
\newblock ISSN 0020-0255.
\newblock \doi{https://doi.org/10.1016/j.ins.2016.01.039}.

\bibitem[Pao and Takefuji(1992)]{pao1992functional}
Y.-H. Pao and Y.~Takefuji.
\newblock Functional-link net computing: theory, system architecture, and
  functionalities.
\newblock \emph{Computer}, 25\penalty0 (5):\penalty0 76--79, 1992.

\bibitem[Huang et~al.(2006)Huang, Zhu, and Siew]{HUANG2006489}
G.-B. Huang, Q.-Y. Zhu, and C.-K. Siew.
\newblock Extreme learning machine: Theory and applications.
\newblock \emph{Neurocomputing}, 70\penalty0 (1):\penalty0 489--501, 2006.
\newblock ISSN 0925-2312.
\newblock \doi{https://doi.org/10.1016/j.neucom.2005.12.126}.
\newblock Neural Networks.

\bibitem[Chen and Liu(2017)]{chen2017broad}
C.L.P. Chen and Z.~Liu.
\newblock Broad learning system: A new learning paradigm and system without
  going deep.
\newblock In \emph{2017 32nd youth academic annual conference of Chinese
  association of automation (YAC)}, pages 1271--1276. IEEE, 2017.
\newblock \doi{10.1109/YAC.2017.7967609}.

\bibitem[Park and Sandberg(1991)]{park1991universal}
J.~Park and I.W. Sandberg.
\newblock Universal approximation using radial-basis-function networks.
\newblock \emph{Neural Computation}, 3\penalty0 (2):\penalty0 246--257, 1991.
\newblock \doi{10.1162/neco.1991.3.2.246}.

\bibitem[Scarselli and Tsoi(1998)]{scarselli1998universal}
F.~Scarselli and A.C. Tsoi.
\newblock Universal approximation using feedforward neural networks: A survey
  of some existing methods, and some new results.
\newblock \emph{Neural Networks}, 11\penalty0 (1):\penalty0 15--37, 1998.
\newblock \doi{10.1016/S0893-6080(97)00097-X}.

\bibitem[Chen et~al.(2018)Chen, Liu, and Feng]{chen2018universal}
C.L.P. Chen, Z.~Liu, and S.~Feng.
\newblock Universal approximation capability of broad learning system and its
  structural variations.
\newblock \emph{IEEE transactions on neural networks and learning systems},
  30\penalty0 (4):\penalty0 1191--1204, 2018.
\newblock \doi{10.1109/TNNLS.2018.2866622}.

\bibitem[Nelsen and Stuart(2021)]{nelsen_2021}
N.~H. Nelsen and A.~M. Stuart.
\newblock The random feature model for input-output maps between banach spaces.
\newblock \emph{SIAM Journal on Scientific Computing}, 43\penalty0
  (5):\penalty0 A3212--A3243, 2021.
\newblock \doi{10.1137/20M133957X}.

\bibitem[Nelsen and Stuart(2024)]{nelsen_2024}
N.~H. Nelsen and A.~M. Stuart.
\newblock Operator learning using random features: A tool for scientific
  computing.
\newblock \emph{SIAM Review}, 66\penalty0 (3):\penalty0 535--571, 2024.
\newblock \doi{10.1137/24M1648703}.

\bibitem[Bolager et~al.(2024)Bolager, {\v{C}}ukarska, Burak, Monfared, and
  Dietrich]{bolager2024koopman}
E.~L. Bolager, A.~{\v{C}}ukarska, I.~Burak, Z.~Monfared, and F.~Dietrich.
\newblock Koopman-informed recurrent neural networks.
\newblock \emph{arXiv preprint arXiv:2410.23467}, 2024.

\bibitem[Eschw{\'e} and Langer(2004)]{eschwe2004variational}
D.~Eschw{\'e} and M.~Langer.
\newblock Variational principles for eigenvalues of self-adjoint operator
  functions.
\newblock \emph{Integral Equations and Operator Theory}, 49\penalty0
  (3):\penalty0 287--321, 2004.
\newblock \doi{10.1007/s00020-002-1209-5}.

\bibitem[Nüske et~al.(2014)Nüske, Keller, P{\'e}rez-Hern{\'a}ndez, Mey, and
  No{\'e}]{nuske2014variational}
F.~Nüske, B.G. Keller, G.~P{\'e}rez-Hern{\'a}ndez, A.S.J.S. Mey, and
  F.~No{\'e}.
\newblock Variational approach to molecular kinetics.
\newblock \emph{Journal of chemical theory and computation}, 10\penalty0
  (4):\penalty0 1739--1752, 2014.
\newblock \doi{10.1021/ct4009156}.

\bibitem[Sch{\"u}tte and Sarich(2013)]{schutte2013metastability}
C.~Sch{\"u}tte and M.~Sarich.
\newblock \emph{Metastability and {M}arkov state models in molecular dynamics},
  volume~24.
\newblock American Mathematical Soc., 2013.
\newblock URL \url{https://bookstore.ams.org/cln-24}.

\bibitem[Banisch and Koltai(2017)]{banisch2017understanding}
R.~Banisch and P.~Koltai.
\newblock Understanding the geometry of transport: {D}iffusion maps for
  {L}agrangian trajectory data unravel coherent sets.
\newblock \emph{Chaos: An Interdisciplinary Journal of Nonlinear Science},
  27\penalty0 (3), 2017.
\newblock \doi{10.1063/1.4971788}.

\bibitem[Koltai et~al.(2018)Koltai, Wu, No{\'e}, and
  Sch{\"u}tte]{koltai2018optimal}
P.~Koltai, H.~Wu, F.~No{\'e}, and C.~Sch{\"u}tte.
\newblock Optimal data-driven estimation of generalized {M}arkov state models
  for non-equilibrium dynamics.
\newblock \emph{Computation}, 6\penalty0 (1):\penalty0 22, 2018.
\newblock \doi{10.3390/computation6010022}.

\bibitem[Klus et~al.(2018)Klus, N{\"u}ske, Koltai, Wu, Kevrekidis, Sch{\"u}tte,
  and No{\'e}]{klus2018data}
S.~Klus, F.~N{\"u}ske, P.~Koltai, H.~Wu, I.~Kevrekidis, C.~Sch{\"u}tte, and
  F.~No{\'e}.
\newblock Data-driven model reduction and transfer operator approximation.
\newblock \emph{Journal of Nonlinear Science}, 28\penalty0 (3):\penalty0
  985--1010, 2018.
\newblock \doi{10.1007/s00332-017-9437-7}.

\bibitem[Mollenhauer et~al.(2020)Mollenhauer, Schuster, Klus, and
  Sch\"utte]{MSKS20}
M.~Mollenhauer, I.~Schuster, S.~Klus, and C.~Sch\"utte.
\newblock Singular value decomposition of operators on reproducing kernel
  {H}ilbert spaces.
\newblock In \emph{Advances in Dynamics, Optimization and Computation}, pages
  109--131, Cham, 2020. Springer.
\newblock \doi{10.1007/978-3-030-51264-4_5}.

\bibitem[Klus et~al.(2019)Klus, Husic, Mollenhauer, and
  No{\'e}]{klus2019kernel}
S.~Klus, B.E. Husic, M.~Mollenhauer, and F.~No{\'e}.
\newblock Kernel methods for detecting coherent structures in dynamical data.
\newblock \emph{Chaos: An Interdisciplinary Journal of Nonlinear Science},
  29\penalty0 (12), 2019.
\newblock \doi{10.1063/1.5100267}.

\bibitem[Petersen et~al.(2012)]{petersen2008matrix}
K.~B. Petersen et~al.
\newblock The matrix cookbook.
\newblock \emph{Technical University of Denmark}, 7\penalty0 (15):\penalty0
  510, 2012.
\newblock URL \url{http://www2.imm.dtu.dk/pubdb/p.php?3274}.

\bibitem[Golub et~al.(1999)Golub, Hansen, and O'Leary]{golub1999tikhonov}
G.~H. Golub, P.~C. Hansen, and D.~P. O'Leary.
\newblock Tikhonov regularization and total least squares.
\newblock \emph{SIAM journal on matrix analysis and applications}, 21\penalty0
  (1):\penalty0 185--194, 1999.

\bibitem[Zhang et~al.(2021)Zhang, Sahai, and Marzouk]{zhang2021computing}
B.~J. Zhang, T.~Sahai, and Y.~M. Marzouk.
\newblock Computing eigenfunctions of the multidimensional
  {O}rnstein--{U}hlenbeck operator.
\newblock \emph{arXiv preprint arXiv:2110.09229}, 2021.

\bibitem[Tabish(2026)]{tabish_thesis_2026}
M.~Tabish.
\newblock \emph{Scientific machine learning for the analysis of complex
  dynamical systems}.
\newblock PhD thesis, Heriot--Watt University and University of Edinburgh,
  2026.

\bibitem[Sch{\"u}tte et~al.(2023)Sch{\"u}tte, Klus, and
  Hartmann]{Schuette_Klus_Hartmann_2023}
C.~Sch{\"u}tte, S.~Klus, and C.~Hartmann.
\newblock Overcoming the timescale barrier in molecular dynamics: {T}ransfer
  operators, variational principles and machine learning.
\newblock \emph{Acta Numerica}, 32:\penalty0 517--673, 2023.
\newblock \doi{10.1017/S0962492923000016}.

\bibitem[Dill et~al.(2008)Dill, Ozkan, Shell, and Weikl]{dill2008protein}
K.~A. Dill, S.~B. Ozkan, M.~S. Shell, and T.~R. Weikl.
\newblock The protein folding problem.
\newblock \emph{Annu. Rev. Biophys.}, 37:\penalty0 289--316, 2008.
\newblock \doi{10.1146/annurev.biophys.37.092707.153558}.

\bibitem[Lindorff-Larsen et~al.(2011)Lindorff-Larsen, Piana, Dror, and
  Shaw]{lindorff2011fast}
K.~Lindorff-Larsen, S.~Piana, R.~O. Dror, and D.~E. Shaw.
\newblock How fast-folding proteins fold.
\newblock \emph{Science}, 334\penalty0 (6055):\penalty0 517--520, 2011.
\newblock \doi{10.1126/science.1208351}.

\bibitem[Arpit et~al.(2019)Arpit, Campos, and Bengio]{arpit_2019}
D.~Arpit, V.~Campos, and Y.~Bengio.
\newblock How to initialize your network? robust initialization for
  {W}eight{N}orm \& {R}es{N}ets.
\newblock In H.~Wallach, H.~Larochelle, A.~Beygelzimer, F.~d\textquotesingle
  Alch\'{e}-Buc, E.~Fox, and R.~Garnett, editors, \emph{Advances in Neural
  Information Processing Systems}, volume~32. Curran Associates, Inc., 2019.
\newblock URL
  \url{https://proceedings.neurips.cc/paper_files/paper/2019/file/e520f70ac3930490458892665cda6620-Paper.pdf}.

\bibitem[Mishkin and Matas(2015)]{mishkin2015all}
D.~Mishkin and J.~Matas.
\newblock All you need is a good init.
\newblock \emph{CoRR}, abs/1511.06422, 2015.
\newblock URL \url{https://api.semanticscholar.org/CorpusID:2780493}.

\bibitem[Datta(2020)]{datta2020survey}
L.~Datta.
\newblock A survey on activation functions and their relation with {X}avier and
  {H}e normal initialization.
\newblock \emph{arXiv preprint arXiv:2004.06632}, 2020.

\bibitem[Bolager et~al.(2023)Bolager, Burak, Datar, Sun, and
  Dietrich]{bolager_2023}
E.~L. Bolager, I.~Burak, C.~Datar, Q.~Sun, and F.~Dietrich.
\newblock Sampling weights of deep neural networks.
\newblock In A.~Oh, T.~Naumann, A.~Globerson, K.~Saenko, M.~Hardt, and
  S.~Levine, editors, \emph{Advances in Neural Information Processing Systems},
  volume~36, pages 63075--63116. Curran Associates, Inc., 2023.
\newblock URL
  \url{https://proceedings.neurips.cc/paper_files/paper/2023/file/c7201deff8d507a8fe2e86d34094e154-Paper-Conference.pdf}.

\bibitem[Rahimi and Recht(2007)]{rahimi2007random}
A.~Rahimi and B.~Recht.
\newblock Random features for large-scale kernel machines.
\newblock \emph{Advances in Neural Information Processing Systems}, 20, 2007.
\newblock URL
  \url{https://proceedings.neurips.cc/paper_files/paper/2007/file/013a006f03dbc5392effeb8f18fda755-Paper.pdf}.

\bibitem[N{\"u}ske and Klus(2023)]{nuske2023efficient}
F.~N{\"u}ske and S.~Klus.
\newblock Efficient approximation of molecular kinetics using random {F}ourier
  features.
\newblock \emph{The Journal of Chemical Physics}, 159\penalty0 (7), 2023.
\newblock \doi{10.1063/5.0162619}.

\bibitem[Fletcher(2000)]{fletcher2000practical}
R.~Fletcher.
\newblock \emph{Practical methods of optimization}.
\newblock John Wiley \& Sons, 2000.

\bibitem[Hotelling(1936)]{hotelling1936cca}
H.~Hotelling.
\newblock Relations between two sets of variates.
\newblock \emph{Biometrika}, 28\penalty0 (3-4):\penalty0 321--377, 12 1936.
\newblock \doi{10.1093/biomet/28.3-4.321}.

\bibitem[Bradbury et~al.(2018)Bradbury, Frostig, Hawkins, Johnson, Leary,
  Maclaurin, Necula, Paszke, Vander{P}las, Wanderman-{M}ilne, and
  Zhang]{jax2018github}
J.~Bradbury, R.~Frostig, P.~Hawkins, M.~J. Johnson, C.~Leary, D.~Maclaurin,
  G.~Necula, A.~Paszke, J.~Vander{P}las, S.~Wanderman-{M}ilne, and Q.~Zhang.
\newblock {JAX}: composable transformations of {P}ython+{N}um{P}y programs,
  2018.
\newblock URL \url{http://github.com/google/jax}.

\bibitem[Heek et~al.(2024)Heek, Levskaya, Oliver, Ritter, Rondepierre, Steiner,
  and van {Z}ee]{flax2020github}
J.~Heek, A.~Levskaya, A.~Oliver, M.~Ritter, B.~Rondepierre, A.~Steiner, and
  M.~van {Z}ee.
\newblock {F}lax: A neural network library and ecosystem for {JAX}, 2024.
\newblock URL \url{http://github.com/google/flax}.

\bibitem[Bittracher et~al.(2018)Bittracher, Koltai, Klus, Banisch, Dellnitz,
  and Sch{\"u}tte]{bittracher2018transition}
A.~Bittracher, P.~Koltai, S.~Klus, R.~Banisch, M.~Dellnitz, and C.~Sch{\"u}tte.
\newblock Transition manifolds of complex metastable systems: Theory and
  data-driven computation of effective dynamics.
\newblock \emph{Journal of nonlinear science}, 28\penalty0 (2):\penalty0
  471--512, 2018.
\newblock \doi{10.1007/s00332-017-9415-0}.

\bibitem[Froyland(2013)]{froyland13}
G.~Froyland.
\newblock An analytic framework for identifying finite-time coherent sets in
  time-dependent dynamical systems.
\newblock \emph{Physica D: Nonlinear Phenomena}, 250:\penalty0 1--19, 2013.
\newblock \doi{10.1016/j.physd.2013.01.013}.

\bibitem[Rypina et~al.(2007)Rypina, Brown, Beron-Vera, Ko{\c{c}}ak, Olascoaga,
  and Udovydchenkov]{rypina2007lagrangian}
I.~I. Rypina, M.~G. Brown, F.~J. Beron-Vera, H.~Ko{\c{c}}ak, M.~J. Olascoaga,
  and I.~A. Udovydchenkov.
\newblock On the {L}agrangian dynamics of atmospheric zonal jets and the
  permeability of the stratospheric polar vortex.
\newblock \emph{Journal of the Atmospheric Sciences}, 64\penalty0
  (10):\penalty0 3595--3610, 2007.
\newblock \doi{10.1175/JAS4036.1}.

\bibitem[Hoffmann et~al.(2021)Hoffmann, Scherer, Hempel, Mardt, de~Silva,
  Husic, Klus, Wu, Kutz, Brunton, et~al.]{hoffmann2021deeptime}
M.~Hoffmann, M.~Scherer, T.~Hempel, A.~Mardt, B.~de~Silva, B.E. Husic, S.~Klus,
  H.~Wu, N.~Kutz, S.L. Brunton, et~al.
\newblock Deeptime: a {P}ython library for machine learning dynamical models
  from time series data.
\newblock \emph{Machine Learning: Science and Technology}, 3\penalty0
  (1):\penalty0 015009, 2021.
\newblock \doi{10.1088/2632-2153/ac3de0}.

\bibitem[Frazier(2018)]{frazier2018tutorialbayesianoptimization}
P.~I. Frazier.
\newblock A tutorial on {B}ayesian optimization, 2018.
\newblock URL \url{https://arxiv.org/abs/1807.02811}.

\end{thebibliography}

%%%%%%%%%%%%%%%%%%%%%%%%%%%%%%%%%%%%%%%%%%%%%%%%%%%%%%%%%%%%
\newpage

\appendix
\renewcommand{\thesection}{\Roman{section}}

\section{Data-driven approximation of self-adjoint operators}

\label{appendix:approx_self_adjoint}

We will use the variational principle to compute eigenfunctions of the linear operator $\mathcal{A}$. Consider a set of $n$ fixed linearly independent basis functions $\psi_i \in \mathcal{D}(\mathcal{A})$, with $i = 1, 2, \dots, n$, and let $ V = \mspan \{ \psi_1, \ldots, \psi_n \} $ so that $V$ is an $n$-dimensional subspace of $ \mathcal{D}(\mathcal{A}) $. We are interested in computing the Galerkin projection of $\mathcal{A}$ onto $V$. A typical choice of basis functions is to consider monomials, radial basis functions, or indicator functions. However, the optimal choice of basis functions depends on the operator. Finding suitable dictionaries is in general an open problem. An arbitrary function $f \in V$ can be written as a linear combination of the form
\begin{align*}
    f(x) = \sum_{i=1}^n w_i\psi_i(x) = w^\top \psi(x),
\end{align*}
where $w = [w_1, \dots, w_n]^\top $ and $\psi \colon \mathbb{R}^d \rightarrow \mathbb{R}^n$ is defined by $\psi(x) = [\psi_1(x), \psi_2(x), \dots, \psi_n(x)]^\top$.
We can use the Rayleigh variational principle to obtain approximations of the eigenvalues $\lambda_i$ and eigenfunctions $\varphi_i$ of~$\mathcal{A}$. For an arbitrary function $f$, we have
\begin{equation*}
    \langle f, \mathcal{A}f \rangle = \left\langle \sum_{i=1}^n w_i \psi_i, \sum_{j=1}^n w_j\mathcal{A}\psi_j \right\rangle
    = \sum_{i=1}^n\sum_{j=1}^n w_i w_j \langle \psi_i, \mathcal{A} \psi_j \rangle
    = w^\top C_{01} w,
\end{equation*}
where $[C_{01}]_{ij} = \langle \psi_i, \mathcal{A} \psi_j \rangle$. Similarly, $\langle f, f \rangle = w^\top C_{00} w$, with  $ [C_{00}]_{ij} = \langle \psi_i, \psi_j \rangle$. That is, we obtain
\begin{equation*}
    \underset{\substack{f}}{\max} \frac{\langle f, \mathcal{A} f \rangle}{\langle f, f \rangle} = \underset{\substack{w \in \mathbb{R}^n}}{\max} \frac{w^\top C_{01} w}{w^\top C_{00} w}.
\end{equation*}
If, on the other hand, we want to compute the smallest eigenvalues and corresponding eigenfunctions, we have to turn the maximization problem into a minimization problem. Since the above quotient is invariant under the scaling of $w$ by some scalar $\gamma \in \mathbb{R}$, we can write this as a constrained optimization problem of the form
\begin{equation*}
    \underset{\substack{w \in \mathbb{R}^n}}{\max}  \quad w^\top C_{01} w  \quad \text{s.t. } \quad w^\top C_{00} w = 1.
\end{equation*}
We can now use the method of Lagrange multipliers \cite{fletcher2000practical} and define the Lagrangian
\begin{align*}
    \mathcal{L}(w, \lambda) &= w^\top C_{01} \ts w - \lambda w^\top C_{00} \ts w.
\end{align*}
As $\mathcal{A}$ is a self-adjoint operator, $C_{01}$ is symmetric so that
\begin{equation*}
    \nabla_w \mathcal{L}(w, \lambda) = (C_{01} + C_{01}^\top) \ts w - 2 \ts \lambda \ts C_{00} \ts w = 2 \ts C_{01} \ts w - 2 \ts \lambda \ts C_{00} \ts w.
\end{equation*}
We thus obtain a generalized eigenvalue problem of the form
\begin{equation*}
    C_{01} \ts w = \lambda \ts C_{00} \ts w.
\end{equation*}
The cost function is maximized by the eigenvector corresponding to the largest eigenvalue of this generalized eigenvalue problem. In fact, the matrix $ A = C_{00}^{-1}C_{01} $ is a representation of the operator $ \mathcal{A} $ projected onto the subspace $ V $, denoted by $ \mathcal{A}_\psi $, which is defined by
\begin{align*}
    \mathcal{A}_{\psi}f(x) := (A c)^\top \psi(x),
\end{align*}
see, e.g., \cite{KKS16}. The eigenfunctions of $\mathcal{A}_{\psi}$ can then be computed using the eigenvectors of $A$. Let $w^{(i)}$ be the $i$th eigenvector of $A$ corresponding to the eigenvalue $\lambda_i$, i.e., $ A \ts w^{(i)} = \lambda_i \ts w^{(i)} $, then defining $\varphi_i(x) = (w^{(i)})^{\top} \psi(x)$, we have
\begin{align}\label{eq:eigfuncs_projected_operator}
   \mathcal{A}_{\psi}\varphi_i(x) = (A w^{(i)})^\top \psi(x) = \lambda_i (w^{(i)})^\top\psi(x) = \lambda_i \ts \varphi_i(x).
\end{align}
Hence, $\varphi_i$ is an eigenfunction of the projected operator. Now suppose we have training data of the form $ \big\{(\psi(x_i), \mathcal{A} \psi(x_i))\big\}_{i=1}^m$. A data-driven approximation of the matrices $C_{00}$ and $C_{01}$ is then given by
\begin{align*}
    \widehat{C}_{00} &= \frac{1}{m}\sum_{i=1}^m \psi(x_i)\psi(x_i)^\top = \frac{1}{m}\Psi_0 \Psi_0^\top, \\
    \widehat{C}_{01} &= \frac{1}{m}\sum_{i=1}^m \psi(x_i)(\mathcal{A} \psi (x_i))^\top = \frac{1}{m}\Psi_0 \Psi_1^\top,
\end{align*}
where
\begin{equation} \label{eq:transformed_matrices}
\begin{split}
    \Psi_0 &=
    \begin{bmatrix}
        \psi(x_{1}) & \psi(x_{2}) & \dots & \psi(x_{m})
    \end{bmatrix}
    \in \mathbb{R}^{n \times m}, \\
    \Psi_1 &=
    \begin{bmatrix}
        \mathcal{A} \psi (x_1) & \mathcal{A} \psi(x_2) & \dots & \mathcal{A} \psi(x_m)
    \end{bmatrix} \in \mathbb{R}^{n \times m}.
\end{split}
\end{equation}
Hence, the final data-driven approximation of the matrix representation of the projected operator $\mathcal{A}_{\psi}$ is given by $\widehat{A} = \widehat{C}_{00}^{-1}\widehat{C}_{01}$. In the transfer operator context, this approach is called \emph{extended dynamic mode decomposition} (EDMD) \cite{williams2015data, KKS16}, which, for reversible systems, is equivalent to the \emph{variational approach of conformation dynamics} (VAC) \cite{nuske2014variational} as shown in \cite{klus2018data}. However, the approach can also be used to compute the ground state and excited states of quantum systems by computing eigenfunctions of the time-independent Sch\"{o}dinger equation. %Numerical results will be presented in Section~\ref{sec:numerical_experiments}.

\section{Data-driven approximation of non-self-adjoint operators}
\label{appendix:approx_non_self_adjoint}

For non-self-adjoint operators $ \mathcal{A} $, we define two sets of $n$ linearly independent basis functions $\{\psi_i\}_{i=1}^n \subset H_1$ and $\{\psi_i'\}_{i=1}^n \subset H_2$ and consider
\begin{align*}
    f(x) = \sum_{i=1}^n w_i \ts \psi_i(x) = w^\top \psi(x) \quad \text{and} \quad
    g(x) = \sum_{i=1}^n w'_i \ts \psi_i'(x) = w'^\top \psi'(x),
\end{align*}
where the vectors $w = [w_1, \dots, w_n]^\top $ and $w' = [w'_1, \dots, w'_n]^\top $ contain the coefficients and $\psi$ and $\psi'$ are vector-valued functions defined by the dictionaries. This results in
\begin{align*}
    \underset{f, g}{\max} \frac{\langle g, \mathcal{A} f \rangle_{H_2}}{\sqrt{\langle f, f \rangle_{H_1} \langle g, g \rangle_{H_2}}} = \underset{w, w' \in \mathbb{R}^n}{\max} \frac{w^\top C_{01} w'}{\sqrt{(w^\top C_{00} w) (w'^\top C_{11} w')}}, \\
\end{align*}
where $[C_{01}]_{ij} = \langle \psi_j', \mathcal{A} \psi_i \rangle_{H_2}$, $[C_{00}]_{ij} = \langle \psi_i, \psi_j \rangle_{H_1}$, and $[C_{11}]_{ij} = \langle \psi_i', \psi_j' \rangle_{H_2}$. Thus, we obtain the constrained optimization problem
\begin{align*}
    \max_{w, w' \in \mathbb{R}^{n}}& \quad w^\top C_{01} w', \\
    \text{s.t.}& \quad \;\, w^\top C_{00} \ts w = 1, \\
    & \quad w'^\top C_{11} \ts w' = 1.
\end{align*}
The Lagrangian is given by
\begin{align*}
    \mathcal{L}(w, w', \lambda, \lambda') &= (w^\top C_{01} w')
    - \frac{\lambda}{2} (w^\top C_{00} \ts w - 1)  - \frac{\lambda'}{2} (w'^\top C_{11} \ts w' - 1).
\end{align*}
The gradients of $\mathcal{L}$ w.r.t.\ $ w $ and $ w' $ are
\begin{align*}
    \nabla_{w} \mathcal{L} &= C_{01} \ts w' - \lambda \ts C_{00} \ts w, \\
    \nabla_{w'} \mathcal{L} &= C_{10} \ts w - \lambda' \ts C_{11} \ts w',
\end{align*}
where $ C_{10} = C_{01}^\top $. Using the optimality conditions, we get
\begin{align}
    C_{01} \ts w' &= \lambda \ts C_{00} \ts w, \label{evp_non_self_1}  \\
    C_{10} \ts w &= \lambda' \ts C_{11} \ts w'. \label{evp_non_self_2}
\end{align}
Multiplying \eqref{evp_non_self_1} from the left by $ w $ and \eqref{evp_non_self_2} by $ w' $ implies that $ \lambda = \lambda' $. Thus, solving \eqref{evp_non_self_2} for $ w' $ and plugging the expression into \eqref{evp_non_self_1}, we have
\begin{equation*}
    C_{00}^{-1} C_{01} C_{11}^{-1} C_{10} \ts w = \lambda^2 w.
\end{equation*}
In practice, we again estimate the matrices from data, i.e., we use the empirical estimates $\widehat{C}_{00}^{-1}$, $\widehat{C}_{01}$, $\widehat{C}_{11}^{-1}$, and $\widehat{C}_{10}$ for the approximation of singular values and singular functions. The derived problem formulation is closely related to \emph{canonical correlation analysis} (CCA) and also VAMPnets, see \cite{mardt2018vampnets, klus2019kernel}.

\begin{remark}
A class of loss functions for computing singular values and functions of the Koopman operator, known as VAMP-$r$ scores, has been derived in~\cite{wu2020variational}. It was shown in \cite{klus2019kernel} that the loss function for $ r = 2 $ is closely related to \emph{canonical correlation analysis} (CCA) \cite{hotelling1936cca} and its kernel and deep learning counterparts.
\end{remark}

\section{RaNNDy variants}
\label{appendix:ranndy_variants}

In Figure \ref{fig:proposed_network}, the two variants of RaNNDy are shown. In the first variant, the output layer represents the optimal basis functions, whereas the output layer in the second variant directly approximates the eigenfunctions using RFM as the basis.

\begin{figure}[H]
    % First Network
    \begin{tcolorbox}[colback=blue!5, colframe=blue!75!black,
    boxrule=1.5pt, arc=5mm, width=0.48\textwidth, enlarge left by=1mm, enlarge right by=2mm,nobeforeafter]
    \centering \textbf{Optimal basis approximator} \\[0.3em]
    \begin{tikzpicture}[rotate=270, scale=0.5, transform shape,
        neuron/.style={circle, draw, minimum size=0.5cm},
        input neuron/.style={neuron, fill=cyan!70!black},
        hidden neuron/.style={neuron, fill=blue!60},
        output neuron/.style={neuron, fill=orange!60},
        every path/.style={->, >=latex},
        xscale=1.2,
        yscale=1.2
    ]

    %neurons
    \def\nInput{3}
    \def\nHidden{6}
    \def\nOutput{4}

    %Input Layer
    \foreach \i in {1,...,\nInput} {
        \node[input neuron] (I\i) at (0,-\i-1.2) {};
    }

    %Hidden Layer
    \foreach \i in {1,...,\nHidden} {
        \node[hidden neuron] (H\i) at (3,-\i+0.4) {};
    }

    % Output Layer
    \foreach \i in {1,...,\nOutput} {
        \node[output neuron] (O\i) at (6,-\i-0.75) {};
    }

    % Input to Hidden Connections
    \foreach \i in {1,...,\nInput} {
        \foreach \j in {1,...,\nHidden} {
            \draw[black!70] (I\i) -- (H\j);
        }
    }

    % Hidden to Output Connections
    \foreach \i in {1,...,\nHidden} {
        \foreach \j in {1,...,\nOutput} {
            \draw[black!70] (H\i) -- (O\j);
        }
    }

    % Layer Labels
    \node[align=center,font=\bfseries, rotate=90] at (0.0, -0.8) {Input \\ layer};
    \node[align=center,font=\bfseries, rotate=90, text=blue] at (3.2, 1.5) {Hidden layers \\ as nonlinear \\ transformation\\
    $R(x_i)$};
    \node[align=center,font=\bfseries, rotate=90, text=black!10!orange] at (6.0, 0.3) {Output \\ layer as \\ basis functions};

    %X_t
    \node[rotate=90] at (-1.8, -3.2) {$x_i \in \R^d$};
    \draw[->, thick] (-1.5, -3.2) -- (-0.5, -3.2);

    % psi(X_t)
    \node[rotate=90, text=black!10!orange] at (8.1, -3.2) {$\psi(x_i) = \sigma(W_o \ts R(x_i)) \in \R^n$};
    \draw[->, thick] (6.5, -3.2) -- (7.5, -3.2);
    \end{tikzpicture}
    \end{tcolorbox}
    \hspace*{-2ex}
    % Second Network
    \begin{tcolorbox}[colback=yellow!6, colframe=blue!75!black,
                      boxrule=1.5pt, arc=5mm, width=0.48\textwidth, enlarge left by=1mm, enlarge right by=2mm,nobeforeafter]
    \centering \textbf{Eigenfunction approximator} \\[0.3em]
    \begin{tikzpicture}[rotate=270, scale=0.5, transform shape,
        neuron/.style={circle, draw, minimum size=0.5cm},
        input neuron/.style={neuron, fill=cyan!70!black},
        hidden neuron/.style={neuron, fill=orange!60},
        output neuron/.style={neuron, fill=black!40!green},
        every path/.style={->, >=latex},
        xscale=1.2,
        yscale=1.2
    ]

    %neurons
    \def\nInput{3}
    \def\nHidden{6}
    \def\nOutput{4}

    %Input Layer
    \foreach \i in {1,...,\nInput} {
        \node[input neuron] (I\i) at (0,-\i-1.2) {};
    }

    %Hidden Layer
    \foreach \i in {1,...,\nHidden} {
        \node[hidden neuron] (H\i) at (3,-\i+0.4) {};
    }

    % Output Layer
    \foreach \i in {1,...,\nOutput} {
        \node[output neuron] (O\i) at (6,-\i-0.75) {};
    }

    % Input to Hidden Connections
    \foreach \i in {1,...,\nInput} {
        \foreach \j in {1,...,\nHidden} {
            \draw[black!70] (I\i) -- (H\j);
        }
    }

    % Hidden to Output Connections
    \foreach \i in {1,...,\nHidden} {
        \foreach \j in {1,...,\nOutput} {
            \draw[black!70] (H\i) -- (O\j);
        }
    }

    % Layer Labels
    \node[align=center,font=\bfseries, rotate=90] at (0.0, -0.8) {Input \\ layer};
    \node[align=center,font=\bfseries, rotate=90, text=orange] at (3.2, 1.5) {Hidden layers \\ as randomized \\ basis functions \\
    $\psi(x_i) = R(x_i)$};
    \node[align=center,font=\bfseries, rotate=90, text=black!50!green] at (6.0, 0.3) {Output \\ layer as \\ eigenfunctions};

    %X_t
    \node[rotate=90] at (-1.8, -3.2) {$x_i \in \R^d$};
    \draw[->, thick] (-1.5, -3.2) -- (-0.5, -3.2);

    % varphi(X_t)
    \node[rotate=90, text=black!50!green] at (8.1, -3.2) {$\varphi(x_i) \in \R^n$};
    \draw[->, thick] (6.5, -3.2) -- (7.5, -3.2);

    \end{tikzpicture}
    \end{tcolorbox}
    \caption{Proposed network architectures: optimal basis approximator (left) and eigenfunction approximator (right).}
    \label{fig:proposed_network}
\end{figure}

\section{Further numerical results}

\subsection{Overdamped Langevin equation}
\label{appendix:overdamped_sde}
The experiments are run on the CPU of a laptop with 12th Gen Intel Core i7 and 32 GB RAM. We use the Python libraries JAX/Flax \cite{jax2018github, flax2020github} for the implementation of the neural networks. Here, we will show the approximation of the eigenvalues and eigenfunctions of Koopman operators associated with simple stochastic differential equations. Consider the overdamped Langevin equation, given by
\begin{equation} \label{eq:overdamp_sde}
    \mathrm{d}\mathbf{X}_t = -\nabla V(\mathbf{X}_t) \ts \mathrm{d}t + \sigma(\mathbf{X}_t) \ts \mathrm{d}W_t,
\end{equation}
where $V(x)$ is the potential of the system, $\sigma(X) = \sqrt{2 \beta^{-1}}$ the diffusion term,  $\beta>0$ the inverse temperature, and $\{W_t\}_{t \geq 0}$ a Wiener process. To generate the required trajectory data $\{x_i, y_i\}_{i=1}^m$, we use the Euler--Maruyama scheme
\begin{equation}\label{eq:euler_maruyama}
    \mathbf{X}_{k+1} = \mathbf{X}_k - h \nabla V(\mathbf{X}_k) + 2 \beta^{-1} \Delta W_k,
\end{equation}
where $h$ is the step size and $\Delta W_k = W_k - W_{k-1} \sim \mathcal{N}(0, h)$, i.e., normally distributed with mean $0$ and variance $h$. We apply RaNNDy, described in Algorithm \ref{alg:eigenfunctions_approximation}, and VAMPnets using $100$ epochs. In order to ensure the comparison is fair, we select the same network architecture consisting of three dense hidden layers comprising 256, 512, and 256 neurons.
%We then approximate the first few dominant eigenfunctions of the Koopman operator for each of the problems below.

The OU process is a one-dimensional system with potential $V(x) = \frac{\alpha}{2} x^2$, where $\alpha$ is the friction coefficient. We choose $\alpha = 1$ and $ \beta = 4$ and generate $m = 20000$ data points with lag time $\tau = 0.5$ by simulating the SDE \eqref{eq:overdamp_sde} using \eqref{eq:euler_maruyama}. The analytically computed eigenvalues and eigenfunctions of the Koopman operator are given by
\begin{equation*}
    \lambda_i = e^{-\alpha(i-1)\tau}, \quad \varphi_i(x) = \frac{1}{\sqrt{(i-1)!}}H_{i-1}\big(\sqrt{\alpha \beta} x \big), \quad i = 1,2,3,\dots,
\end{equation*}
where $H_i$ denotes the $i$th probabilists' Hermite polynomial.

%The numerically approximated eigenvalues and eigenfunctions are shown in Figures \ref{subfig:ou_eigvals} and \ref{subfig:ou_eigfuncs}, respectively. The numerical approximations computed by RaNNDy and VAMPnets match the analytical solutions. Since we create the data using a single long trajectory containing very few data points near the boundaries of the domain, the approximations are less accurate in these regions.

\begin{figure}[t]
\centering

\begin{minipage}{0.06\textwidth}
    \centering
    \rotatebox{90}{\Large VAMP score}
\end{minipage}
\hfill
\begin{minipage}{0.9\textwidth}
    \centering

    \subfloat[\label{subfig:ou2dcomp_normal}]
    {\includegraphics[width=0.48\linewidth]{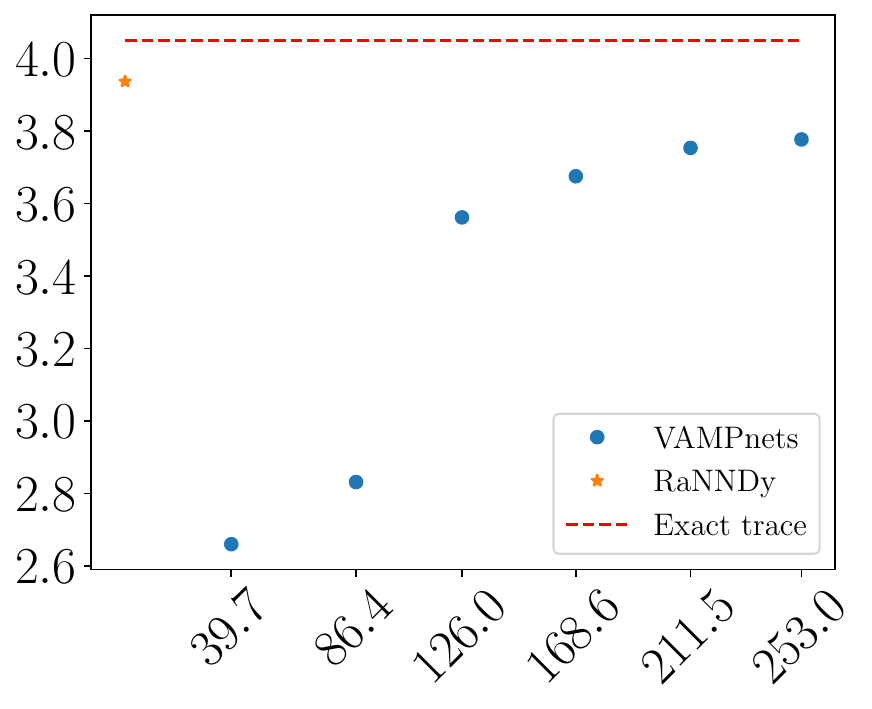}}
    \hfill
    \subfloat[\label{subfig:ou2dcomp_uniform}]
    {\includegraphics[width=0.48\linewidth]{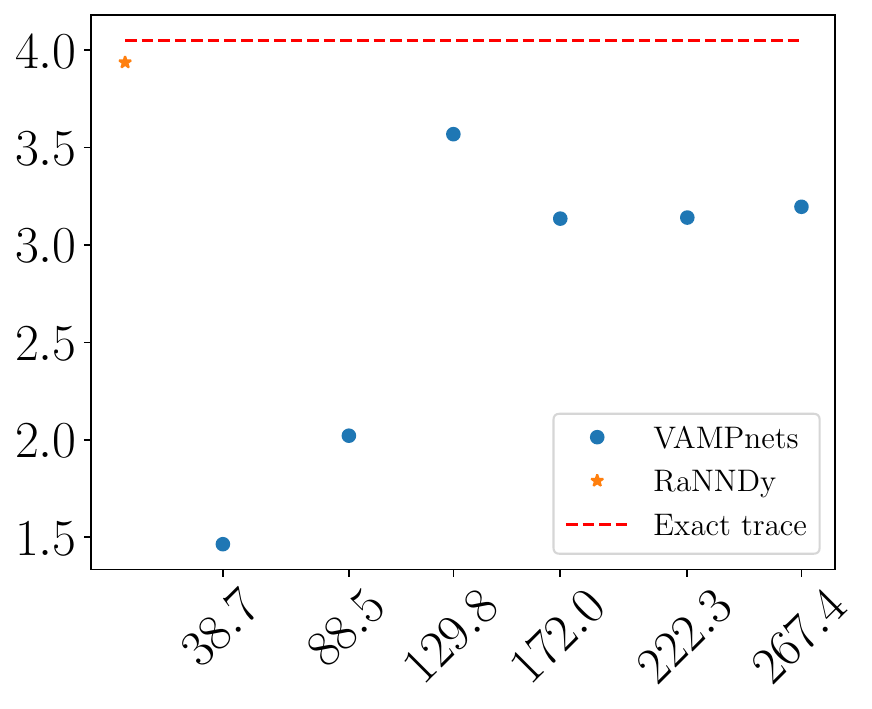}}

    \vspace{0.3cm}

    \subfloat[\label{subfig:ou2dcomp_lecun_normal}]
    {\includegraphics[width=0.48\linewidth]{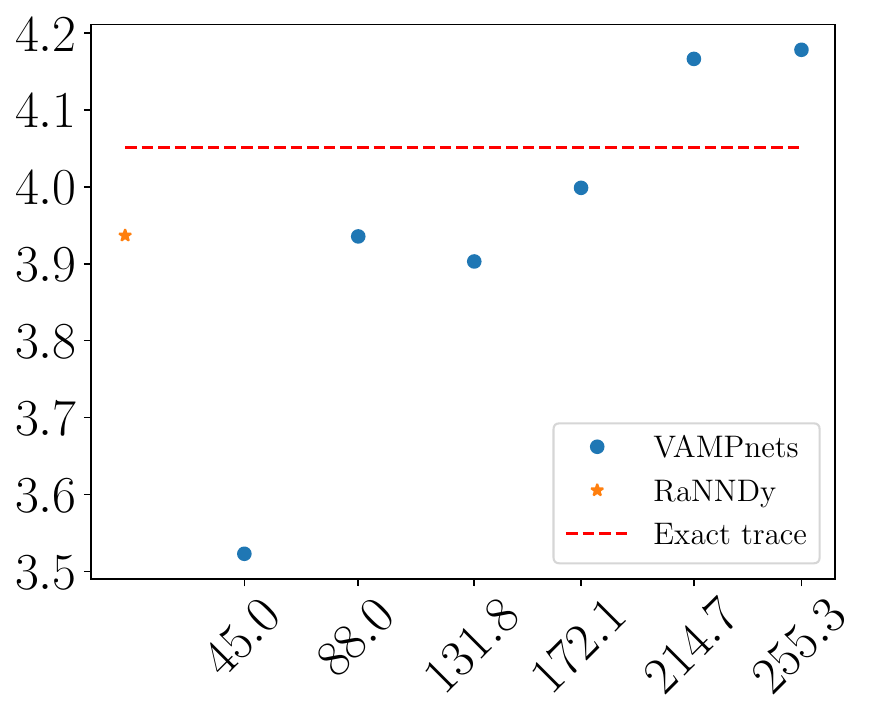}}
    \hfill
    \subfloat[\label{subfig:ou2dcomp_normal_poor}]
    {\includegraphics[width=0.48\linewidth]{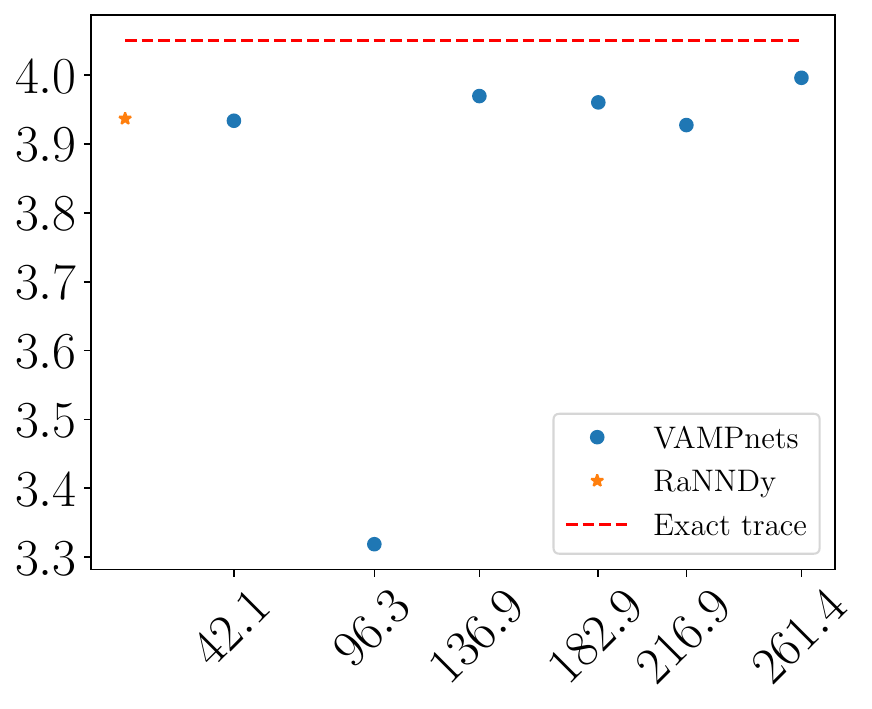}}

    \vspace{0.4cm}

    % {\Large Epochs (time in sec.)}

\end{minipage}
\caption{RaNNDy and VAMPnet comparison for the approximation of the sum of five dominant eigenvalues of a 2D OU process. The run times for VAMPnet, on the x-axis, correspond to 20, 40, 60, 80, 100, and 120 epochs, respectively. The weights in VAMPnets are initialized using (a) a normal distribution, (b) a uniform distribution, (c) the LeCun normal distribution (which is also the default initialization in Flax). The overestimation of the true VAMP score in (c) might be due to ill-conditioned matrices or spectral pollution. (d) Sometimes a poor initialization of the weights leads to fluctuations during training.}
\label{fig:multidim_ou}
\end{figure}

In Figure \ref{fig:multidim_ou}, we provide further analysis illustrating the robustness of RaNNDy for the approximation of transfer operators. The eigenvalues and eigenfunctions associated with multi-dimensional OU processes can also be computed analytically \cite{zhang2021computing, tabish_thesis_2026}. We compute the sum of the five largest eigenvalues using RaNNDy and VAMPnets for a 2D OU process. The results show that VAMPnets sometimes converge slowly, whereas RaNNDy typically performs well, provided that the weights are initialized properly. It is important to note that RaNNDy also gives poor approximations for poor initializations of weights.

% \begin{figure}
%     \centering
%     \begin{minipage}[t]{0.31\textwidth}
%         \subfloat[][\label{subfig:hidden_convergence_oudim4}]{\includegraphics[width=0.9\textwidth]{pics/hidden_convergence_dim4OU.pdf}}
%     \end{minipage}
%     \begin{minipage}[t]{0.31\textwidth}
%         \subfloat[][\label{subfig:hidden_convergence_chig}]{\includegraphics[width=0.9\textwidth]{pics/hidden_convergence_chig.pdf}
%     }
%     \end{minipage}
%     \begin{minipage}[t]{0.31\textwidth}
%         \subfloat[][\label{subfig:hidden_convergence_nug2}]{\includegraphics[width=0.9\textwidth]{pics/hidden_convergence_nug2.pdf}}
%     \end{minipage}
%     \caption{Convergence of RaNNDy with an increasing number of neurons in the hidden layer to approximate dominant eigenvalues of (a) the four-dimensional OU process in (b) chignolin molecule. (c) NuG2 molecule.}
%     \label{fig:hidden_convergence_results}
% \end{figure}

\subsection{Two-dimensional potentials}

\paragraph{Uncertainty quantification: } One of the advantages of our framework is that it can be used to quantify the uncertainty in the approximation of the eigenfunctions or singular functions. The performance of a single randomized neural network can be unstable due to the random nature of features in the hidden layers. A standard way to make these models more robust is to utilize an ensemble learning approach \cite{malik2023random}. In ensemble learning, multiple models (base learners) are created to solve a specific task. The base learners are then combined to produce an output of the aggregated model. The base learners can be created by selecting multiple random features from the data and averaging the output of each model to return the final output. For the eigenfunction approximation, we create the base learners in Algorithm \ref{alg:eigenfunctions_approximation} by sampling multiple weights and biases $(W, b)$ from a distribution. Each random feature set gives us a random basis via hidden layers to approximate the eigenfunctions. Hence, we can approximate the operator's eigenfunctions using Algorithm \ref{alg:eigenfunctions_approximation} by creating multiple models and calculating the average eigenfunction for all models and quantify uncertainty by calculating the variances across the models. We will show in the numerical results section that, as expected, the uncertainty is high in regions containing only a few data points and low in densely sampled regions.

\begin{figure}
    \centering
    \begin{minipage}[t]{0.44\textwidth}
        \subfloat[][\label{subfig:ls2d_potential}]{\includegraphics[height=5cm]{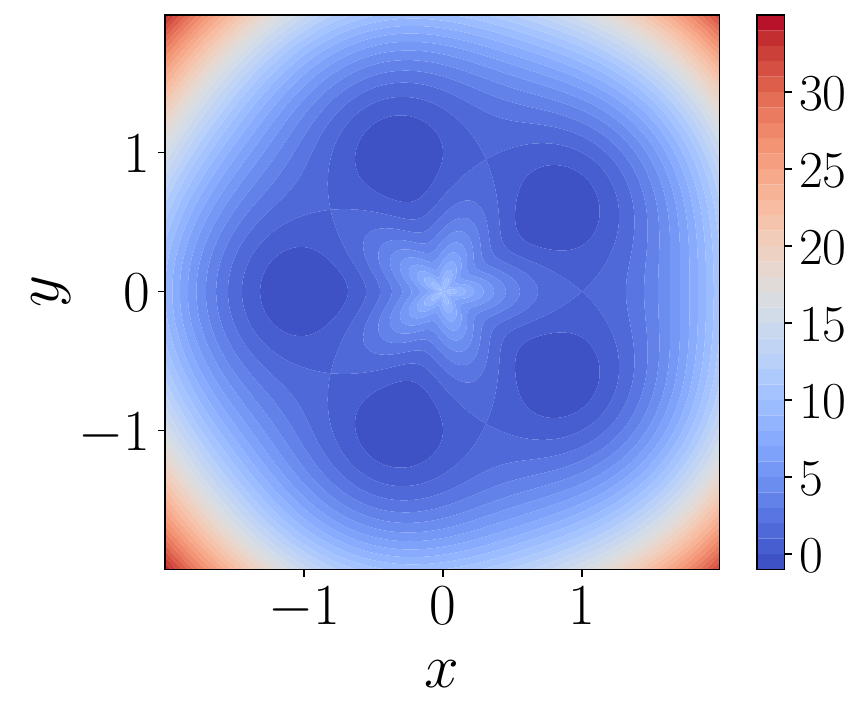}}
    \end{minipage}
    \begin{minipage}[t]{0.44\textwidth}
        \subfloat[][\label{subfig:ls2d_eigf_cluster}]{\includegraphics[height=5cm, width=0.8\textwidth]{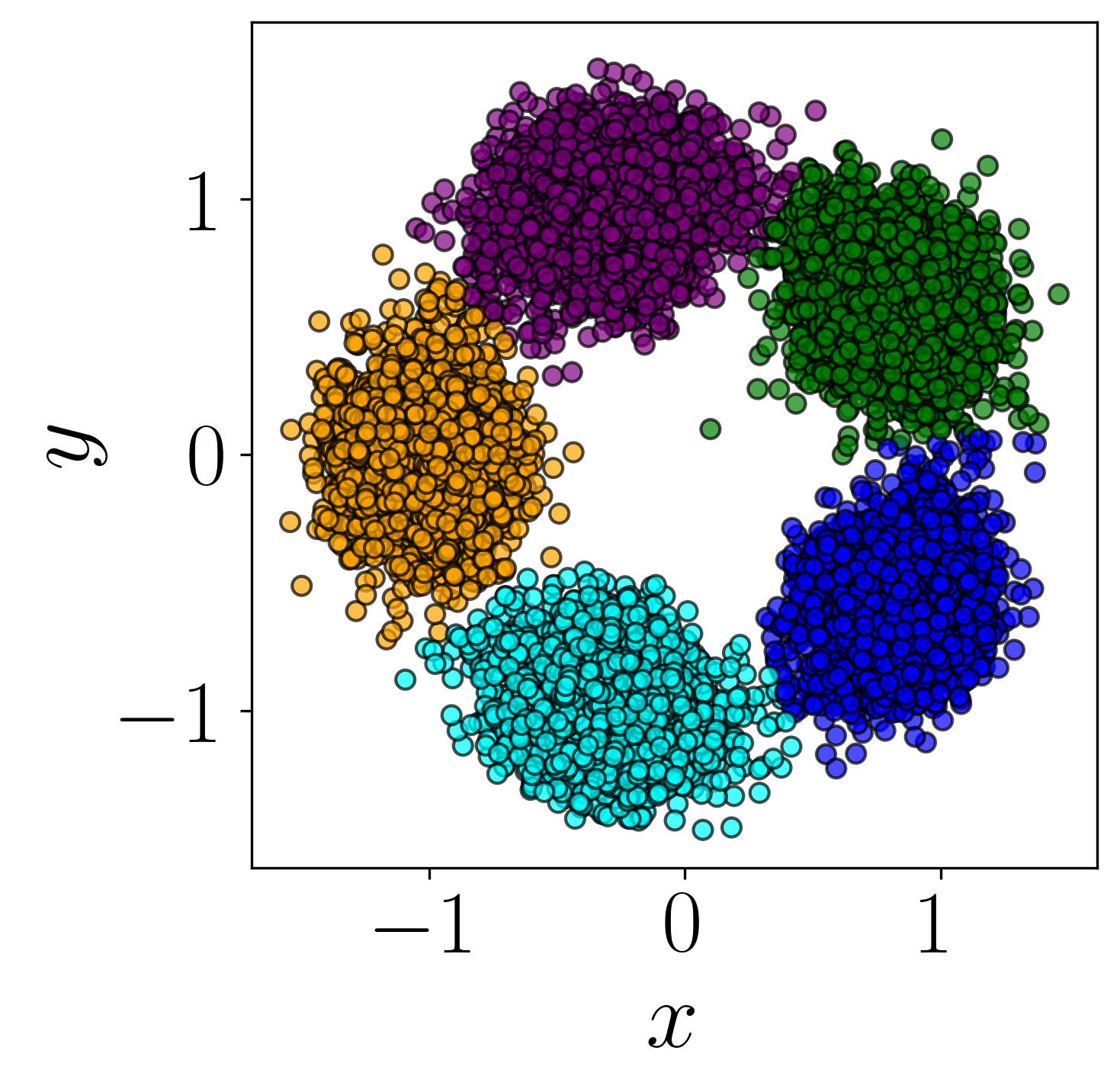}}
    \end{minipage}\\
    \begin{minipage}[t]{0.44\textwidth}
        \subfloat[][\label{subfig:tw2d_potential}]{\includegraphics[width=0.9\textwidth]{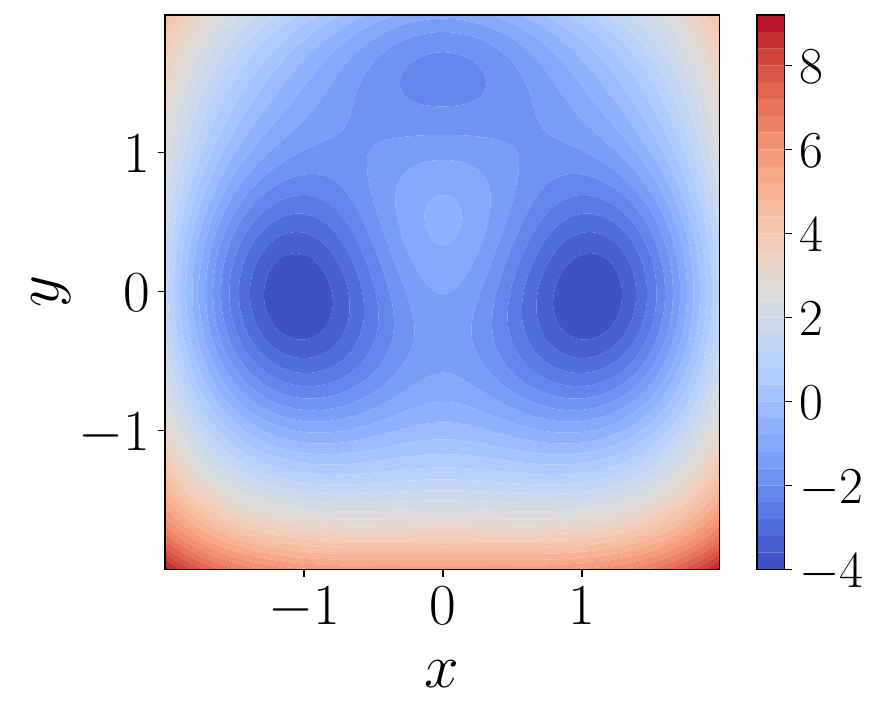}}
    \end{minipage}
    \begin{minipage}[t]{0.44\textwidth}
        \subfloat[][\label{subfig:tw2d_eigvals_uncer}]{\includegraphics[width=1.\textwidth]{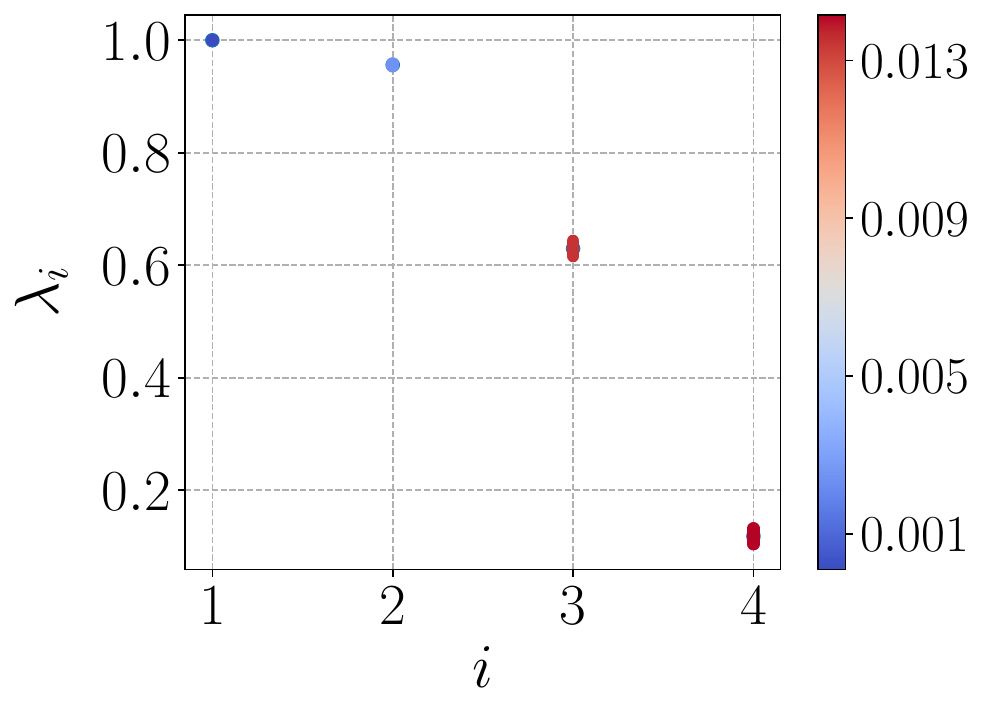}}
    \end{minipage}\\
    \hspace{1.0cm}\begin{minipage}[t]{0.4\textwidth}
        \subfloat[][\label{subfig:tw2d_second_eigf}]{\includegraphics[width=0.9\textwidth]{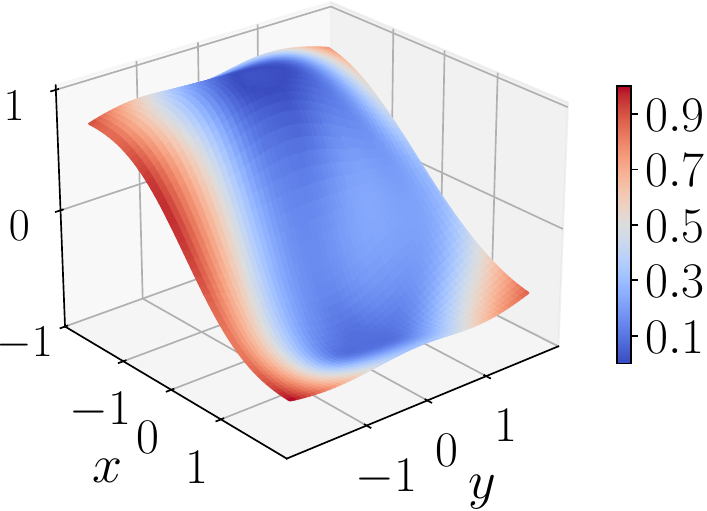}}
    \end{minipage}
    \hspace{1.0cm}\begin{minipage}[t]{0.4\textwidth}
        \subfloat[][\label{subfig:tw2d_third_eigf}]{\includegraphics[width=0.9\textwidth]{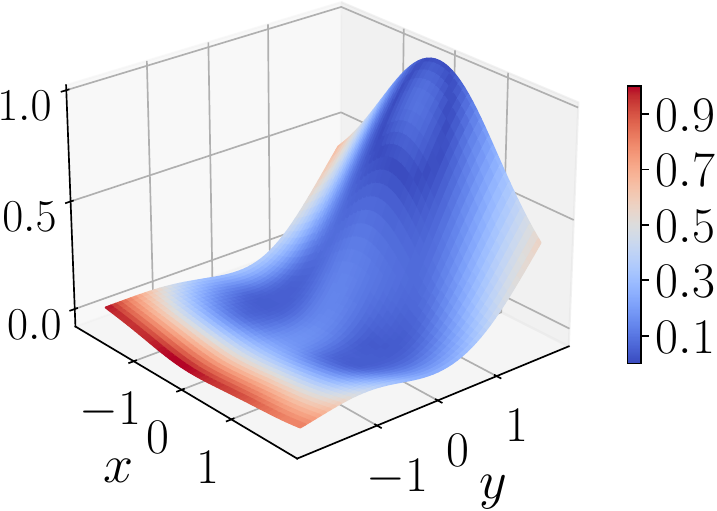}}
    \end{minipage}
    \caption{RaNNDy results for 2D potentials. (a) The lemon-slice potential function with $n=5$ wells. (b) Clustering of the five dominant eigenfunctions into $n=5$ clusters. (c) The triple-well 2D potential. (d) Uncertainty in the dominant eigenvalues.  (e) Second eigenfunction of the Koopman operator colored according to the uncertainties in approximation in different regions. (f) Third dominant eigenfunction of the Koopman operator colored according to the approximation uncertainties.}
    \label{fig:2dpotential_results}
\end{figure}

We now consider the two-dimensional lemon-slice potential
\begin{equation*}
    V(x, y) = \cos\left(q \cdot \mathrm{atan2}(y, x)\right) + 10 \left( \sqrt{x^2 + y^2} - 1 \right)^2,
\end{equation*}
which comprises $q$ wells arranged on a circle \cite{bittracher2018transition}. The potential for $ q = 5 $ is visualized in Figure~\ref{subfig:ls2d_potential}. We also consider the triple-well potential
\begin{align*}
        V(x, y)&=3 \exp({-x^2 - (y - 1/3)^2})-3\exp({-x^2-(y - 5/3)^2}) \\ \nonumber
    & -5 \exp(-(x - 1)^2 - y^2) - 5\exp(-(x + 1)^2 - y^2) \\ \nonumber
    &+\cfrac{2}{10} x^4 + \cfrac{2}{10}(y - 1/3)^4.
\end{align*}
We generate $m=20000$ data points for the above potentials and then approximate the five and three dominant eigenfunctions of the Koopman operator for the lemon-slice and triple-well potentials, respectively. For the lemon-slice potential, the clustering of the dominant eigenfunctions into five clusters is presented in Figure \ref{subfig:ls2d_eigf_cluster}, clearly indicating the five wells of the system. Furthermore, we apply the proposed ensemble learning RaNNDy to quantify the uncertainties in the RaNNDy approximation of the triple-well eigenfunctions by generating $100$ models. Figure \ref{subfig:tw2d_eigvals_uncer} represents the uncertainty in the approximated eigenvalues, showing that the uncertainty increases for smaller eigenvalues, i.e., faster timescales. In Figures \ref{subfig:tw2d_second_eigf} and \ref{subfig:tw2d_third_eigf}, we show the average approximation (for all models) of the second and third dominant eigenfunctions colored according to the uncertainty in the approximation in different regions of the domain. The results illustrate that, as expected, regions with more data points have less uncertainty.

\paragraph{Comparison:} Using the same network architecture, our approach requires significantly less computational time compared to VAMPnets, see Table~\ref{tab:sde_results}. This is due to the fact that RaNNDy does not optimize the weights of all the hidden layers. The obtained accuracy for the benchmark problems, however, is comparable, which implies that the randomly generated features are suitable basis functions. It is important to note that in cases where a randomized basis is not optimal, the VAMPnets approach might provide better approximations because of its flexibility in optimizing the basis.

\begin{table}
    \centering
    \caption{RaNNDy and VAMPnets run times for overdamped Langevin equations.}
    \begin{tabular}{l c c}
        \toprule
         & \multicolumn{2}{c}{Computational time (in sec.)} \\
        \cmidrule{2-3}
        System & RaNNDy & VAMPnets (100 epochs) \\
        \midrule
        OU process & 0.32 & 121.35  \\
        Lemon-slice & 0.33 & 117.65 \\
        Triple-well & 0.65 & 125.70 \\
        \bottomrule
    \end{tabular}
    \label{tab:sde_results}
\end{table}

\subsection{Bickley jet}
\label{appendix:bickley_jet}
\begin{figure}
    \centering
    \subfloat[][\label{subfig:bickley_timestep_0}]{\includegraphics[width=.31\textwidth]{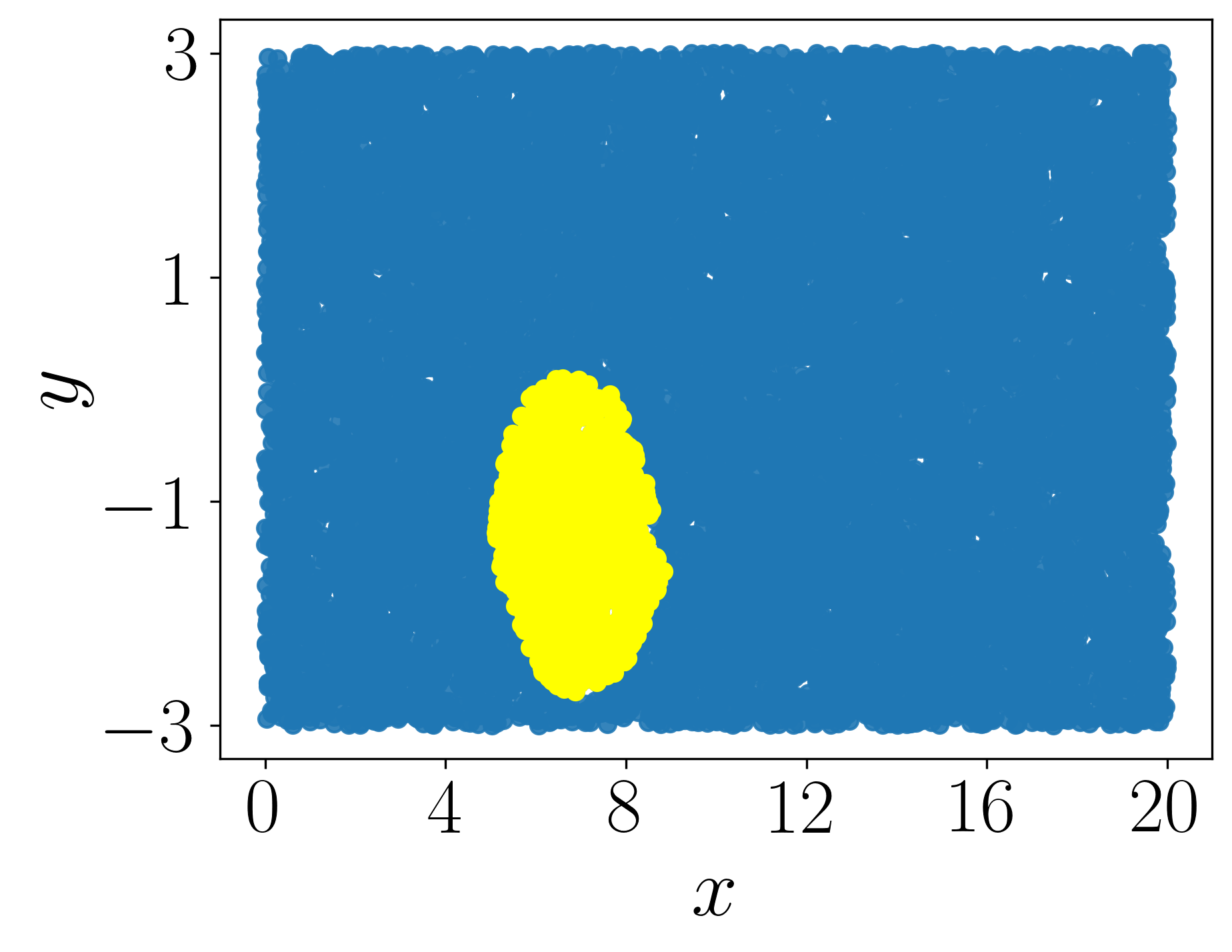}}\quad
    \subfloat[][\label{subfig:bickley_timestep_25}]{\includegraphics[width=.31\textwidth]{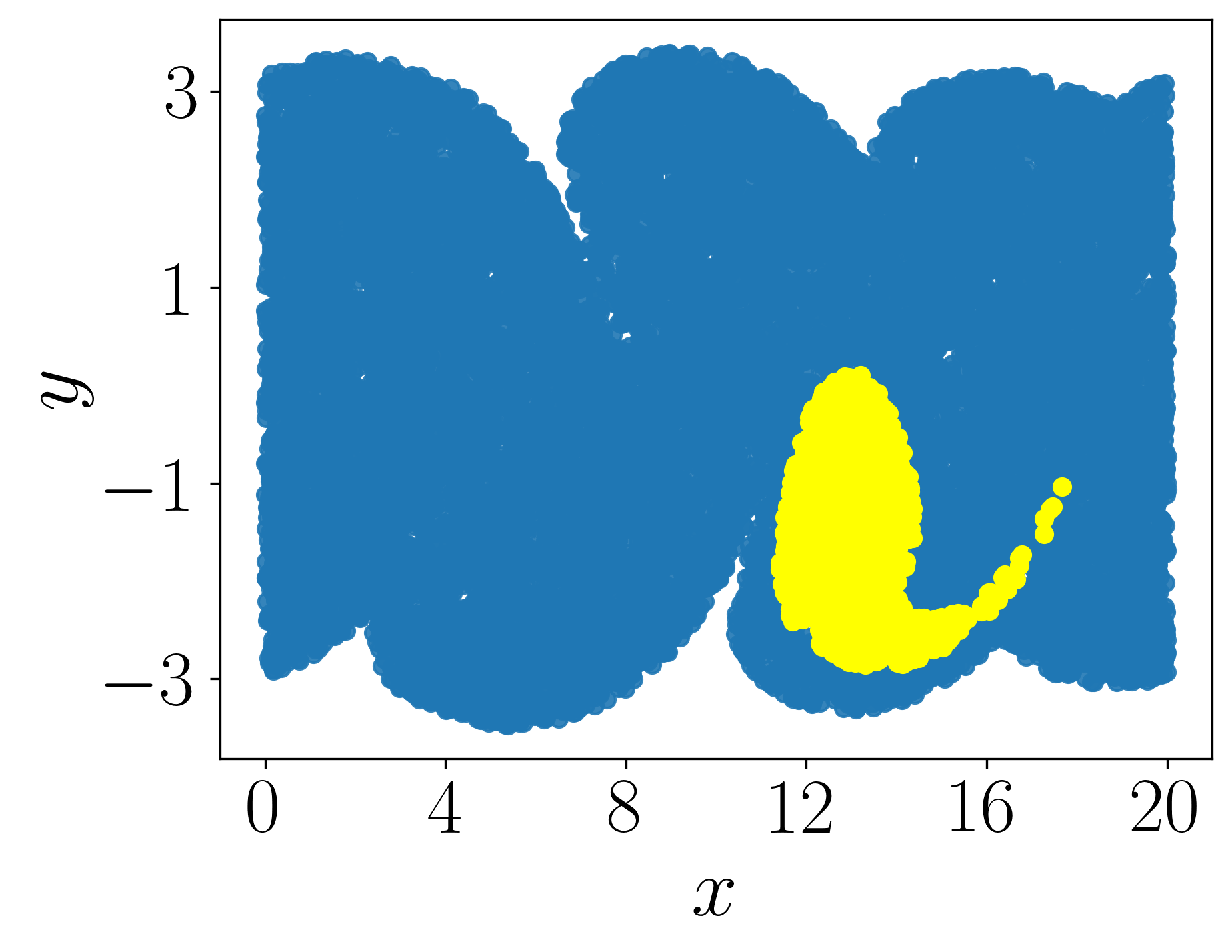}}\quad
    \subfloat[][\label{subfig:bickley_timestep_50}]{\includegraphics[width=.31\textwidth]{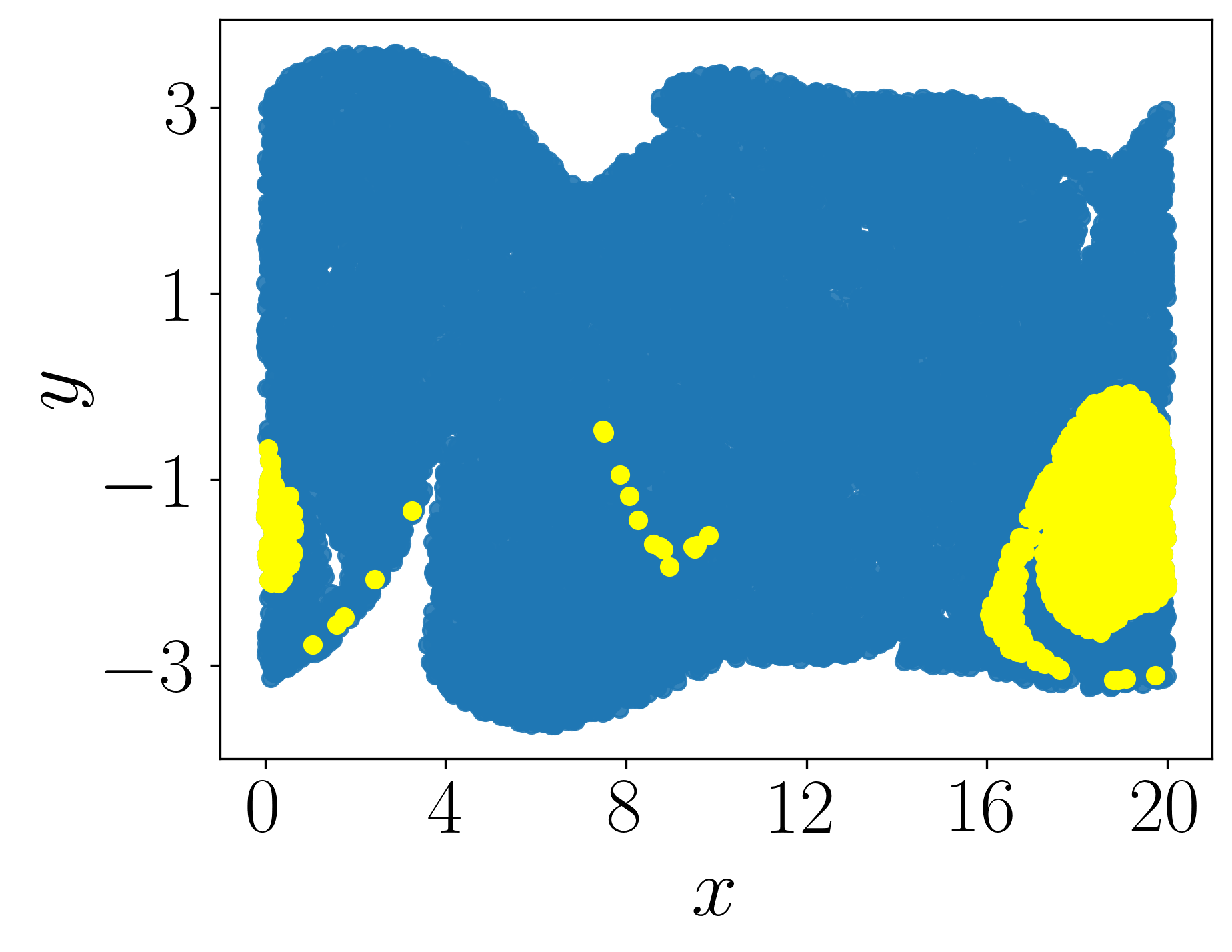}}\quad
    \caption{Bickley jet flow at time (a) $t = 0$, (b) $t = 25$, and (c) $t= 50$. The particles in yellow stay in close proximity during the flow, forming a coherent set.}
    \label{fig:bickley_flow}
\end{figure}

\begin{figure}
    \centering
    \subfloat[][\label{subfig:bickley_eigf1}]{\includegraphics[width=.31\textwidth]{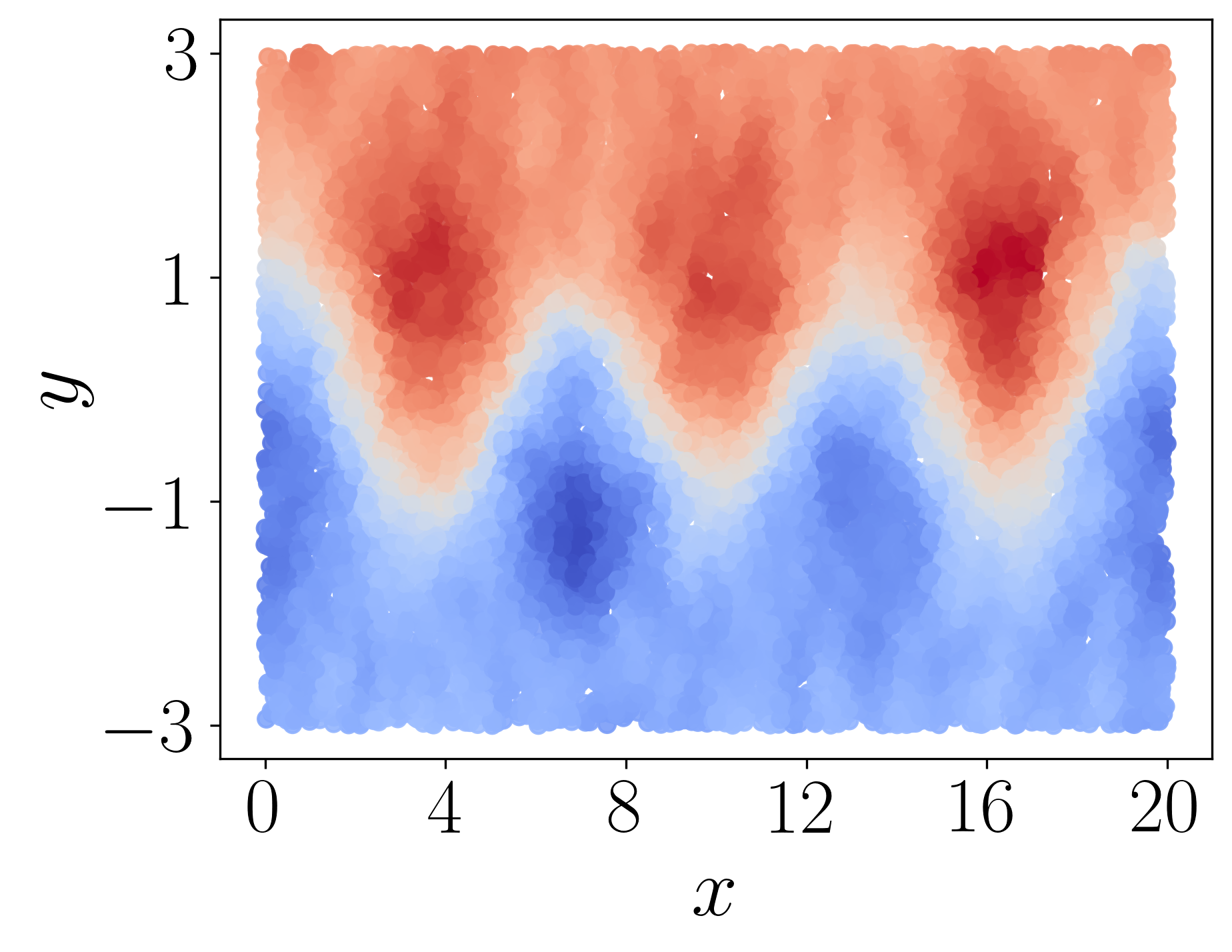}}\quad
    \subfloat[][\label{subfig:bickley_eigf2}]{\includegraphics[width=.31\textwidth]{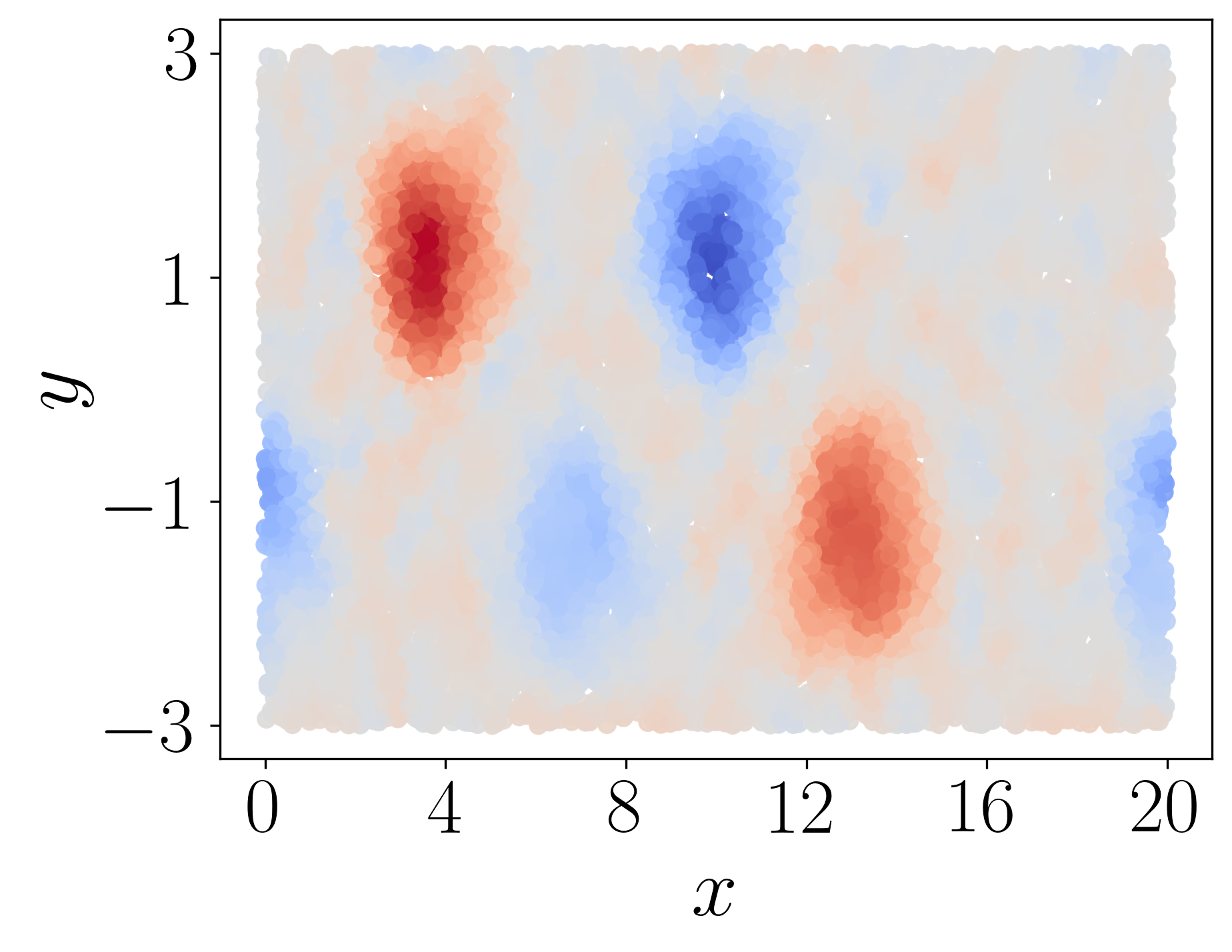}}\quad
    \subfloat[][\label{subfig:bickley_eigf_cluster}]{\includegraphics[width=.31\textwidth]{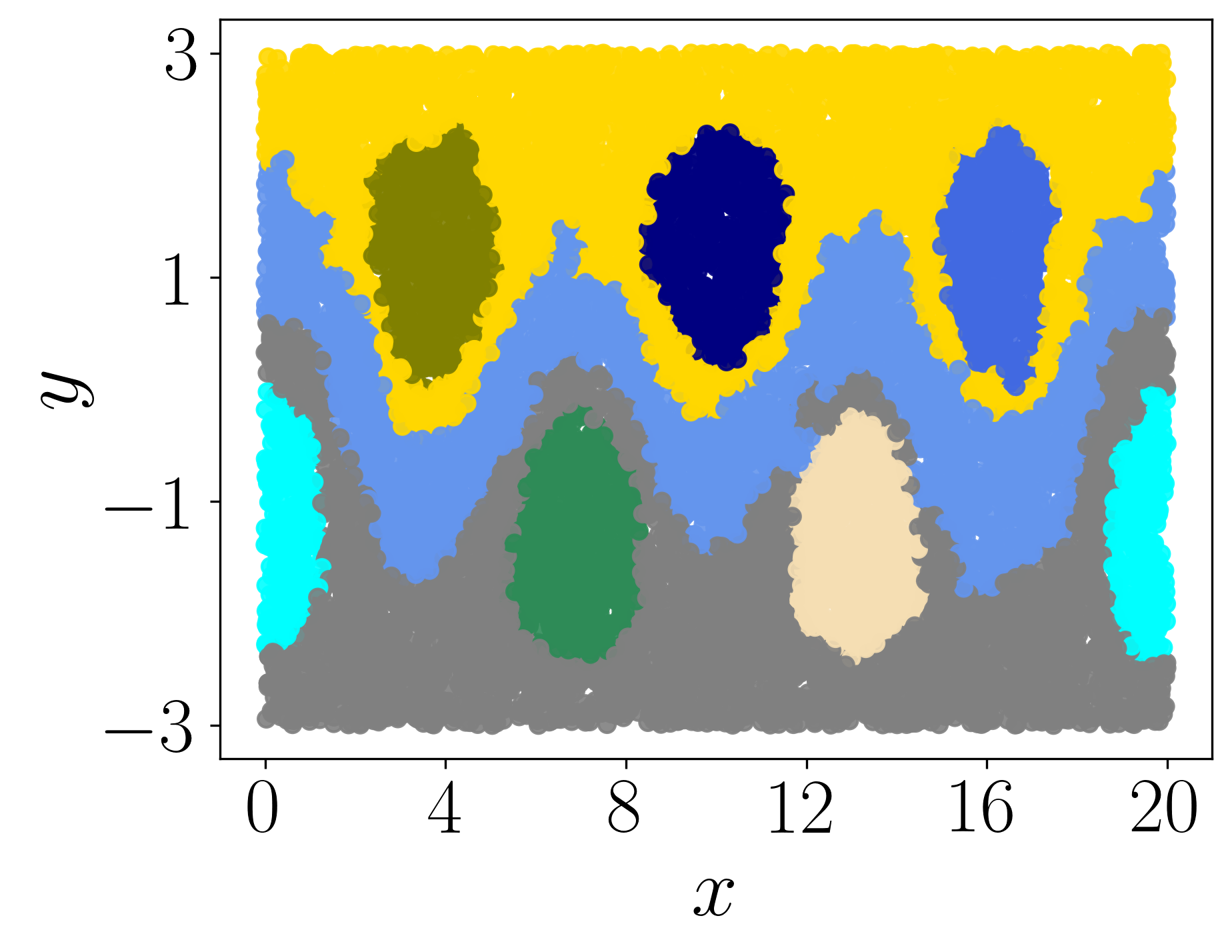}}\quad
    \caption{RaNNDy results for the Bickley jet. (a) First and (b) second dominant singular functions. (c) Clustering of the dominant singular functions into nine clusters.}
    \label{fig:bickley_results}
\end{figure}

We will now illustrate how the proposed framework can be used to detect coherent structures \cite{froyland13, banisch2017understanding}. As a simple benchmark problem, we consider the well-known Bickley jet, which is an example of an idealized stratospheric flow \cite{rypina2007lagrangian}. The dynamics are governed by a system of non-autonomous ordinary differential equations (ODEs).
\begin{equation*}
    \begin{bmatrix}
    \dot{x} \\
    \dot{y}
    \end{bmatrix}
    =
    \begin{bmatrix}
    - \dfrac{\partial \Phi}{\partial y} \\
    \phantom{-} \dfrac{\partial \Phi}{\partial x}
    \end{bmatrix},
\end{equation*}
with the stream function
\begin{align*}
    \Phi(x, y, t) &= c_3 y - U_0 L \tanh\left(\frac{y}{L}\right)
    + A_3 U_0 L \, \text{sech}^2\left(\frac{y}{L}\right) \cos(k_1 x) \\
    & + A_2 U_0 L \, \text{sech}^2\left(\frac{y}{L}\right) \cos(k_2 x - \sigma_2 t) \\
    & + A_1 U_0 L \, \text{sech}^2\left(\frac{y}{L}\right) \cos(k_1 x - \sigma_1 t).
\end{align*}
The data is generated using the methods and parameters from the Python library \emph{deeptime} \cite{hoffmann2021deeptime}. Trajectories of $m=15 \ts 000$ uniformly sampled data points within the domain $ [0, 20] \times [-4, 4] $ are simulated by integrating the points from $t_0 = 0$ to $t_1 = 40$. A few snapshots of the flow at different times $ t $ are visualized in Figure \ref{fig:bickley_flow}. We approximate the dominant nine right singular functions of the operator $ \mathcal{T}^\tau $ and use them to decompose the state space into nine different clusters. The clusters represent the coherent sets in the flow. Two dominant singular functions are shown in Figures~\ref{subfig:bickley_eigf1} and \ref{subfig:bickley_eigf2} and the resulting clusters in Figure \ref{subfig:bickley_eigf_cluster}. Additionally, we provide the video of the coherent sets in the flow identified using RaNNDy in the supplementary material.

\section{Ablation studies}

\label{appendix:ablation_studies}

\subsection{Grid search using the VAMP score}
\label{subsec:grid_search_vamp}

The initialization of a randomized neural network plays an important role in determining the quality and numerical stability of the approximated Koopman operator. As discussed in Section~\ref{sec:ranndy_operator_approximation}, the randomized feature map
\[
R(x) = \sigma(Wx+b)
\]
generates a finite-dimensional feature space for the approximation. Consequently, the distributions used to sample the weights $W$ and biases $b$, as well as their associated parameters, can influence the richness of the resulting feature space, the conditioning of the matrices involved in the approximation, and the quality of the estimated operator. An unsuitable initialization may produce features with insufficient variability or excessive redundancy, potentially leading to poorly conditioned covariance matrices and less reliable estimates. It is therefore useful to investigate the sensitivity of the approximation to the choice of initialization.

We employ a grid-search strategy based on the VAMP score to compare different initialization schemes. Specifically, we consider candidate distributions for the weights and biases and apply Algorithm~\ref{alg:eigenfunctions_approximation} to each combination. We then evaluate the resulting approximation using the VAMP score, defined here as the sum of the dominant eigenvalues of the approximated operator matrix as discussed in Section \ref{sec:ranndy_operator_approximation}. The combination yielding the highest score is selected as the preferred initialization among the candidates considered. This procedure provides a systematic and computationally straightforward means of assessing how different initialization distributions affect the resulting approximation.

In addition to the choice of distribution, we investigate the influence of distributional parameters. For example, when Gaussian distributions are used, we denote the weight and bias scales by $\omega_W$ and $\omega_b$, respectively. The feature map can then be written as
\[
R(x;\omega)
=
\sigma\bigl(W(\omega_W)x+b(\omega_b)\bigr),
\qquad
\omega=(\omega_W,\omega_b),
\]
where $W(\omega_W)$ and $b(\omega_b)$ denote random variables sampled from distributions parameterized by the corresponding scales. A grid search over candidate values of $\omega$ can be used to identify parameter ranges that yield favorable approximation performance. For larger or more complex parameter spaces, Bayesian optimization~\cite{frazier2018tutorialbayesianoptimization} might provide a more suitable potential alternative.

Tables~\ref{tab:grid_search_distributions_ou} and~\ref{tab:grid_search_distributions_chignolin} report the results of these experiments for the Ornstein--Uhlenbeck (OU) process and the Chignolin molecule, respectively. The distributional comparison indicates that the choice of initialization can have a noticeable effect on the VAMP score. In the OU experiment, Gaussian weight and bias initialization achieves a score of $2.1902$, whereas uniform initialization for both weights and biases yields a substantially lower score of $0.4332$. These results suggest that the distribution used for initialization can influence the quality of the resulting finite-dimensional approximation. However, the optimal distribution may depend on the underlying dynamical system.

\begin{table}[ht]
\centering
\caption{Grid search over weight and bias initialization distributions for the OU process using the VAMP score.}
\label{tab:grid_search_distributions_ou}
\begin{tabular}{cccc}
\toprule
Weight Distribution &
Bias Distribution &
Time (s) &
VAMP Score \\
\midrule
normal          & normal  & 0.3186 & 2.1902 \\
normal          & uniform & 0.2529 & 2.1782 \\

lecun\_normal   & normal  & 0.7938 & 2.1018 \\
lecun\_normal   & uniform & 0.8749 & 2.0938 \\

uniform         & normal  & 0.3097 & 0.4270 \\
uniform         & uniform & 0.3197 & 0.4332 \\

glorot\_uniform & normal  & 0.3796 & 1.6128 \\
glorot\_uniform & uniform & 0.2670 & 1.6128 \\
\bottomrule
\end{tabular}
\end{table}

\begin{table}[ht]
\centering
\caption{Grid search over weight and bias initialization distributions for the Chignolin molecule using the VAMP score.}
\label{tab:grid_search_distributions_chignolin}
\begin{tabular}{cccc}
\toprule
Weight Distribution &
Bias Distribution &
Time (s) &
VAMP Score \\
\midrule
normal        & normal  & 1.5467 & 2.7475 \\
normal        & uniform & 0.1705 & 2.7879 \\
normal        & zeros   & 0.6861 & 2.7802 \\

lecun\_normal & normal  & 0.2509 & 2.0616 \\
lecun\_normal & uniform & 0.1862 & 2.0510 \\
lecun\_normal & zeros   & 0.1554 & 2.0708 \\

uniform       & normal  & 0.1793 & 1.0000 \\
uniform       & uniform & 0.1479 & 1.0000 \\
uniform       & zeros   & 0.1929 & 1.0000 \\

glorot\_uniform & normal  & 0.2629 & 1.9793 \\
glorot\_uniform & uniform & 0.2048 & 2.0316 \\
glorot\_uniform & zeros   & 0.1518 & 2.0524 \\
\bottomrule
\end{tabular}
\end{table}

\begin{table}
\centering
\caption{Grid search over weight and bias initialization scales for the OU process.}
\label{tab:grid_search_scales_ou}
\begin{tabular}{cccc}
\toprule
Weight Scale ($\omega_w$) &
Bias Scale ($\omega_b$) &
Time (s) &
VAMP Score \\
\midrule
0.001 & 0.001 & 0.3179 & 1.0000 \\
0.001 & 0.5   & 0.2962 & 1.0000 \\
0.001 & 3.0   & 0.2718 & 1.0000 \\
0.001 & 10.0  & 0.2725 & 1.0000 \\

0.05  & 0.001 & 0.4051 & 1.6128 \\
0.05  & 0.5   & 0.4017 & 1.6128 \\
0.05  & 3.0   & 0.2847 & 1.0000 \\
0.05  & 10.0  & 0.3175 & 1.0000 \\

0.5   & 0.001 & 0.2899 & 1.3881 \\
0.5   & 0.5   & 0.3310 & 2.2060 \\
0.5   & 3.0   & 0.2564 & 2.2009 \\
0.5   & 10.0  & 0.3167 & 2.1764 \\

1.0   & 0.001 & 0.3165 & 1.3013 \\
1.0   & 0.5   & 0.3169 & 2.1973 \\
1.0   & 3.0   & 0.3159 & 2.1955 \\
1.0   & 10.0  & 0.6499 & 2.1883 \\

5.0   & 0.001 & 0.3242 & 0.9650 \\
5.0   & 0.5   & 0.2887 & 2.1786 \\
5.0   & 3.0   & 0.4262 & 2.1931 \\
5.0   & 10.0  & 0.2694 & 2.1972 \\

10.0  & 0.001 & 0.9138 & 0.9430 \\
10.0  & 0.5   & 0.2512 & 2.1611 \\
10.0  & 3.0   & 0.3093 & 2.1820 \\
10.0  & 10.0  & 0.8377 & 2.1812 \\
\bottomrule
\end{tabular}
\end{table}

In Table~\ref{tab:grid_search_scales_ou}, the scale-sensitivity experiment further illustrates that the effect of initialization parameters is not uniform across the parameter space. For example, when both the weight and bias scales are set to $0.001$, the resulting VAMP score is $1.0000$, compared with $2.2060$ when both scales are set to $0.5$. In contrast, several configurations with moderate or larger weight scales and non-negligible bias scales produce scores close to $2.2$.

In practice, standard normal initialization provides a simple and effective default for the examples considered here. Nevertheless, the grid-search results demonstrate that this choice should not be regarded as universally optimal. In particular, the VAMP score provides a useful criterion for comparing candidate initializations, but it should ideally be complemented by diagnostics based on matrix conditioning, etc. For the examples considered in this paper, a standard grid search over different distributions takes around 5--10 seconds depending on the problem size and implementation, see Table \ref{tab:protein_initialization_results} for protein folding results.

\begin{table}[htbp]
    \centering
    \caption{Comparison of initialization distribution for protein molecules}
    \label{tab:protein_initialization_results}
    \begin{tabular}{l c p{5.5cm} c}
        \hline
        System &
        Avg. Search Time (s) &
        Best Initialization &
        Best VAMP Score \\
        \hline
        Chignolin &  5--6 &
        normal weights + normal biases & 2.57 \\

        %Chignolin & 11--12 &
        %Similar values (can be replaced by a few GD steps on scales) & 2.58 \\

        NuG2  & 9--10 &
        LeCun normal weights + Zero biases & 4.61 \\
        \hline
    \end{tabular}
\end{table}

\subsection{Robustness to different random seeds}
\label{subsec:random_seed_robustness}

Having investigated the sensitivity of the approximation to the choice of the distribution used for initialization and its parameters, we next examine the variability arising from different random realizations drawn from a fixed distribution. Although the chosen initialization distribution determines the richness of the randomized features, individual realizations may still lead to different finite-dimensional feature spaces and, consequently, different approximations of the Koopman operator. Assessing this variability is important for determining whether the observed approximation quality is representative of the chosen initialization scheme or depends strongly on a particular random seed.

To quantify this effect, we generate an ensemble of randomized neural networks using independent random seeds while keeping the initialization distribution and the remaining experimental settings fixed. For each realization, we apply Algorithm~\ref{alg:eigenfunctions_approximation} and compute the four dominant eigenvalues for the OU process. We then calculate the ensemble mean and standard deviation of each eigenvalue, as well as the corresponding VAMP score. This procedure allows us to assess both the variability of the individual spectral estimates and the stability of their aggregate score across independent random initializations.

For the OU process, the ensemble-averaged leading eigenvalues are
\begin{align*}
\bar{\lambda}
=
[0.999533,\;0.612335,\;0.352190,\;0.218048].
\end{align*}
The corresponding mean-plus-minus-standard-deviation intervals are
\begin{align*}
\begin{aligned}
\lambda_1 &: [0.999297,\;0.999768],\\
\lambda_2 &: [0.612077,\;0.612592],\\
\lambda_3 &: [0.347320,\;0.357061],\\
\lambda_4 &: [0.213200,\;0.222877].
\end{aligned}
\end{align*}
The first two eigenvalues exhibit particularly small variability across the ensemble, whereas the third and fourth exhibit somewhat larger, but still comparatively limited variation. These results suggest that the leading spectral estimates are reasonably stable under changes in the random seed used for the initialization scheme and experimental configuration considered here.

We further investigate the convergence of the ensemble-averaged VAMP score as the number of realizations increases. Table~\ref{tab:ensemble_convergence} reports the sum of the ensemble-averaged leading four eigenvalues for ensemble sizes ranging from $50$ to $600$. The resulting scores remain close to $2.182$ throughout the experiment, with no systematic increase or decrease as the ensemble size grows. In particular, the score changes from $2.182460$ for an ensemble of $50$ realizations to $2.182425$ for an ensemble of $600$ realizations. The small variation across these ensemble sizes suggests that the sum of estimated eigenvalues is relatively insensitive to the particular collection of random seeds.

These observations thus indicate that, once an appropriate initialization scheme has been selected, the resulting Koopman approximation exhibits limited sensitivity to individual random realizations. However, the degree of robustness may depend on the dynamical system, the feature dimension, the initialization distribution, and the numerical conditioning of the matrices involved.

\begin{table}[ht]
\centering
\caption{Convergence of the ensemble-averaged VAMP score with increasing ensemble size. The reported score is the sum of the average of the leading four eigenvalues over all models in the ensemble.}
\label{tab:ensemble_convergence}
\begin{tabular}{cc}
\toprule
Ensemble Size & Sum of Average Eigenvalues (VAMP Score) \\
\midrule
50  & 2.182460 \\
100 & 2.182810 \\
200 & 2.182152 \\
400 & 2.182144 \\
600 & 2.182425 \\
\bottomrule
\end{tabular}
\end{table}

\subsection{Comparison with random Fourier features (RFF)}

Random Fourier features (RFFs)~\cite{rahimi2007random} provide a randomized approach to constructing finite-dimensional feature representations. They can be viewed as a special case of RaNNDy, corresponding to a single hidden layer with a cosine activation function and randomly sampled frequency vectors. RFFs have already been successfully used for learning Koopman operators~\cite{nuske2023efficient}. It has in particular been shown that dominant eigenfunctions can be efficiently represented using Fourier features.

However, the RaNNDy framework provides a broader family of randomized feature maps, allowing for different choices of activation functions and weight and bias distributions. This flexibility could be particularly relevant when approximating non-self-adjoint Koopman operators, for which the dominant singular functions might not be well represented by a fixed Fourier basis. Thus, although RFFs can be effective for certain classes of dynamical systems, their suitability may depend on the structure of the underlying dynamics.

To investigate this, we use the Bickley jet as a benchmark problem and compare RaNNDy with RFFs, focusing on the approximation of the dominant singular functions of the Koopman operator. For higher bandwidths of the Gaussian kernel and across random feature dimensions of 2048 and 4096, the RFF approach achieves low VAMP scores of 3.98 and 3.50, respectively, which is not sufficient for the Bickley jet; however, with smaller bandwidth, i.e., high scaling of the normal distribution of weights $\sigma$, the VAMP score increases and the accuracy of RFFs and RaNNDy are similar, see Table \ref{tab:rff_ranndy_comparison}. It is important to note that for a high number of random features in RFF, the size of the covariance matrices increases and makes the computations time-consuming, whereas RaNNDy attains a similar accuracy faster using a small number of neurons in the last hidden layer to reduce the size of the approximated matrices. The observed VAMP score is consistent with the expected VAMP score for the Bickley jet, which exhibits nine dominant singular patterns (see the accompanying video and, for example,~\cite{froyland13, hoffmann2021deeptime}). Furthermore, the greater flexibility in allowing different activation functions and distributions of randomized weights and biases may lead to better representations of more complicated structures.

Table \ref{tab:rff_ranndy_comparison} highlights the advantage of the RaNNDy approach, as we use a narrower final hidden layer (architecture [4096, 128]) before solving the variational problem. This reduces the dimensionality of the feature space $N$ and, consequently, the size of the covariance matrices involved in the generalized eigenvalue problem, leading to significantly lower computational costs, while maintaining a similar VAMP score compared to the corresponding random Fourier feature models. The computational complexity of solving the generalized eigenvalue problems in RaNNDy scales cubically with the number of neurons in the last hidden layer of the network $N$ and typically requires $\mathcal{O}(N^3)$ operations. Storing the covariance  and cross-covariance matrices $C_{00}$ and $C_{01}$ requires $\mathcal{O}(N^2)$ memory.

\begin{table}[ht]
\centering
\caption{Comparison of RFFs and RaNNDy for the non-self-adjoint Koopman operator associated with the non-reversible Bickley jet. The feature configuration for RaNNDy specifies the number of neurons in its hidden layers.}
\label{tab:rff_ranndy_comparison}
\begin{tabular}{lcccc}
\toprule
Method &
Feature Configuration & Scale ($\sigma$) &
VAMP Score &
Time (s) \\
\midrule
RFF & 2048 & 1.0 & 3.98 & 6.47 \\
RFF & 4096 & 1.0 & 3.50 & 84.47 \\
RFF & 2048 & 2.0 & 7.25 & 7.72 \\
RFF & 4096 & 2.0 & 7.11 & 72.47 \\
RFF & 8192 & 2.0 & 6.59 & 422.49 \\
RFF & 2048 & 3.0 & 8.17 & 7.28 \\
RFF & 4096 & 3.0 & 8.12 & 69.72 \\
\midrule
RaNNDy (best initialization) & [4096, 256] & 0.5 & 7.72 & 1.63 \\
RaNNDy (scales search) & [4096, 256] & 0.5 & 7.72 & 13.63 \\
\bottomrule
\end{tabular}
\end{table}

\end{document}